\documentclass{article}

\usepackage[final]{neurips_2024}
\usepackage[utf8]{inputenc} % allow utf-8 input
\usepackage[T1]{fontenc}    % use 8-bit T1 fonts
\usepackage{hyperref}       % hyperlinks
\usepackage{url}            % simple URL typesetting
\usepackage{booktabs}       % professional-quality tables
\usepackage{amsfonts}       % blackboard math symbols
\usepackage{nicefrac}       % compact symbols for 1/2, etc.
\usepackage{microtype}      % microtypography
\usepackage{xcolor}         % colors
\usepackage{bibentry}
\usepackage{listings} 
\usepackage{pifont}
\usepackage{graphicx}
\usepackage{wrapfig}
\usepackage{amsmath}
\usepackage{algorithm}
\usepackage{algorithmic}
\usepackage{multirow}
\definecolor{background}{gray}{0.95}
\usepackage[utf8]{inputenc} % 告诉 LaTeX 你的源文件是 UTF-8 编码
\usepackage[T1]{fontenc}      % 使用更现代的字体编码，以获得更好的输出效果
\usepackage{booktabs} % 用于专业的表格线条 (\toprule, \midrule, \bottomrule)
\usepackage{array}    % 用于定义新的列类型 (如此处的 M 类型)
\usepackage{makecell} % 用于在单元格内实现换行 (\makecell)
\usepackage{caption} % <--- 添加此行以定义 \DeclareCaptionStyle
\usepackage{float}   % <--- 添加此行以定义 \floatstyle 和 \newfloat
\usepackage{enumitem}
\usepackage{pgfplots} % 主要的绘图宏包
\pgfplotsset{compat=1.18} % 设置版本兼容性，推荐使用较新版本
\usepackage{booktabs} % 用于绘制高质量表格的横线

\usepackage{threeparttable}

\colorlet{punct}{red!60!black}
\definecolor{background}{HTML}{F5F5F5} % 定义背景色
\definecolor{string}{rgb}{0.58,0,0.82}  % 定义字符串颜色 (深紫色)
\definecolor{key}{rgb}{0.0,0.5,0.0}     % 定义键颜色 (深绿色)

\DeclareCaptionStyle{ruled}{labelfont=normalfont,labelsep=colon,strut=off} 
\floatstyle{ruled}
\newfloat{listing}{tb}{lst}{}
\floatname{listing}{Listing}
\newcolumntype{M}[1]{>{\centering\arraybackslash}m{#1}}
\colorlet{punct}{red!60!black}
\definecolor{background}{HTML}{F5F5F5}
\definecolor{delim}{RGB}{20,105,176}
\colorlet{numb}{magenta!60!black}

\title{FineState-Bench: A Comprehensive Benchmark for \\ Fine-Grained State Control in GUI Agents}

\author{
  Fengxian Ji\textsuperscript{\rm 1}\thanks{Equal contribution.}, 
  Jingpu Yang\textsuperscript{\rm 1}\footnotemark[1], 
  Zirui Song\textsuperscript{\rm 2}\footnotemark[1], 
  Yuanxi Wang\textsuperscript{\rm 1}, Zhexuan Cui\textsuperscript{\rm 1},\\ \textbf{Yuke Li}\textsuperscript{\rm 1}, \textbf{Qian Jiang}\textsuperscript{\rm 1}, \textbf{Miao Fang}\textsuperscript{\rm 1}, \textbf{Xiuying Chen}\textsuperscript{\rm 2}\\
  \textsuperscript{\rm 1}Northeastern University, China\\
  \textsuperscript{\rm 2}MBZUAI, United Arab Emirates
}

\begin{document}
    \maketitle
    % \footnotetext{$\dagger$ Corresponding author.}
    \begin{abstract}

With the rapid advancement of generative artificial intelligence technology, Graphical User Interface (GUI) agents have demonstrated tremendous potential for autonomously managing daily tasks through natural language instructions. However, current evaluation frameworks for GUI agents suffer from fundamental flaws: existing benchmarks overly focus on coarse-grained task completion while neglecting fine-grained control capabilities crucial for real-world applications. To address this, we introduce \textbf{FineState-Bench}, the first evaluation and diagnostic standard for fine-grained GUI proxy operations, designed to quantify fine-grained control. This multi-platform (desktop, Web, mobile) framework includes 2257 task benchmarks in four components and uses a four-phase indicator for comprehensive perception-to-control assessment. To analyze perception and positioning for refined operations, we developed the plug-and-play Visual Diagnostic Assistant (\textbf{VDA}), enabling the first quantitative decoupling analysis of these capabilities. Experimental results on our benchmark show that the most advanced models achieve only 32.8\% fine-grained interaction accuracy. Using our VDA in controlled experiments, quantifying the impact of visual capabilities, we showed that ideal visual localization boosts Gemini-2.5-Flash's success rate by 14.9\%. Our diagnostic framework confirms for the first time that the primary bottleneck for current GUI proxies is basic visual positioning capability.All resources are fully open-source. 
github: \url{https://github.com/AnonymousThewarehouse/FineState-Bench}
huggingface: \url{https://huggingface.co/datasets/Willtime2006/Static-FineBench}

\end{abstract}

\section{Introduction}

Recent advances in Large Vision-Language Models (LVLMs) have enabled a new class of GUI agents capable of autonomous operation, representing a key frontier in AI (e.g., GUI Grounding\cite{nguyen2024improvedguigroundingiterative}). By integrating visual understanding with language instructions, these agents can interact with complex applications in a human-like manner. Models such as CogAgent \cite{hong2024cogagent} and AppAgent \cite{zhang2025appagent} exemplify this trend, impacting human-computer interaction by completing complex tasks across real-world applications. Consequently, to guide and evaluate progress in this burgeoning field, benchmarks like AITW \cite{rawles2023android} and ScreenSpot~\cite{cheng2024seeclick} have been developed. The establishment of these benchmarks provides a standardized measure for model performance, accelerating model innovation.

However, prevailing benchmarks widely adopt coarse-grained success criteria, such as merely determining the correctness of a UI element's category ~\cite{rawles2023android,cheng2024seeclick}. This approach systematically overlooks the fine-grained manipulation capabilities essential for real-world applications, for instance, precisely selecting a color value or highlighting a specific sentence.
This neglect of fine-grained state awareness and precise control leads to an illusion of capability — a phenomenon where models achieve inflated scores on benchmarks yet underperform on practical, high-precision tasks, We show the difference between general instructions and refined instructions in Figure 1.This constitutes the most critical research gap in current evaluation systems.

\begin{figure}[htbp] % 使用 [htbp] 提供更灵活的浮动位置
    \centering
    \includegraphics[
        width=0.63\textwidth,  
        height=9cm     
    ]{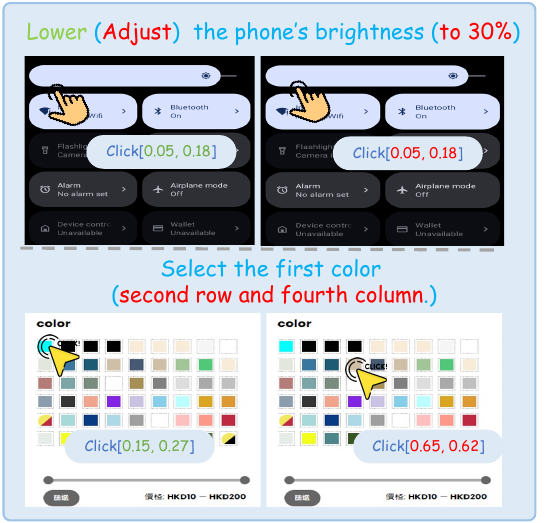}
    \caption{Show the difference between our command instructions and the ordinary task-oriented instructions.}
    \label{fig:demo}
\end{figure}

The prevailing evaluation gap has resulted in divergent explanations for agent task failures. While some research ascribes these failures to poor underlying visual localization abilities~\cite{OS-Atlas,wei2022chainofthought}, other work highlights defects in high-level perception. Such contradictory findings stem from the absence of a standardized diagnostic framework that can quantitatively disentangle the impacts of visual perception errors from those of high-level perception failures on task outcomes. We define this challenge in pinpointing the precise cause of failure as the diagnostic gap, as it obstructs a deeper comprehension of the model's performance and the ability to provide targeted feedback.

To address the Evaluation Gap and Diagnostic Gap, we introduce FineState-Bench, a comprehensive framework for both evaluation and diagnostics. First, it fills the evaluation void with fine-grained tasks across major platforms, shifting the paradigm from a binary task success metric to a quantitative assessment of interaction control precision \cite{lee2024benchmarking, Li2025FerretUI2}. Second, it breaks the diagnostic bottleneck with an innovative framework featuring a plug-and-play VDA. Inspired by the controlled variable method, it allows researchers to quantitatively isolate and measure the contribution of visual grounding failures to task outcomes \cite{zheng2024gpt4v, Chen2024EDGE, dardouri2024visual}. Thus, our twofold contribution is a much-needed benchmark for fine-grained control and a generalizable diagnostic methodology to pinpoint agent capability bottlenecks. In summary, our contributions include:

\begin{itemize}
    \item Identify and quantify the evaluation gap in GUI agent benchmarks, and design reproducible multi-dimensional metrics that expose the real limitations of high-scoring models on fine-grained control tasks.
    \item Establish and open-source FineState-Bench, the first benchmark for fine-grained state awareness and control, spanning three platforms and four component types for comprehensive evaluation.
    \item Introduce a plug-and-play VDA to enable quantitative diagnosis of visual grounding bottlenecks, facilitating deeper understanding of capability limitations in GUI agents.
\end{itemize}

\begin{figure*}[htbp]
    \centering
    \includegraphics[
        width=1.0\textwidth,      
        height=0.4\textheight,    
        % keepaspectratio         
    ]{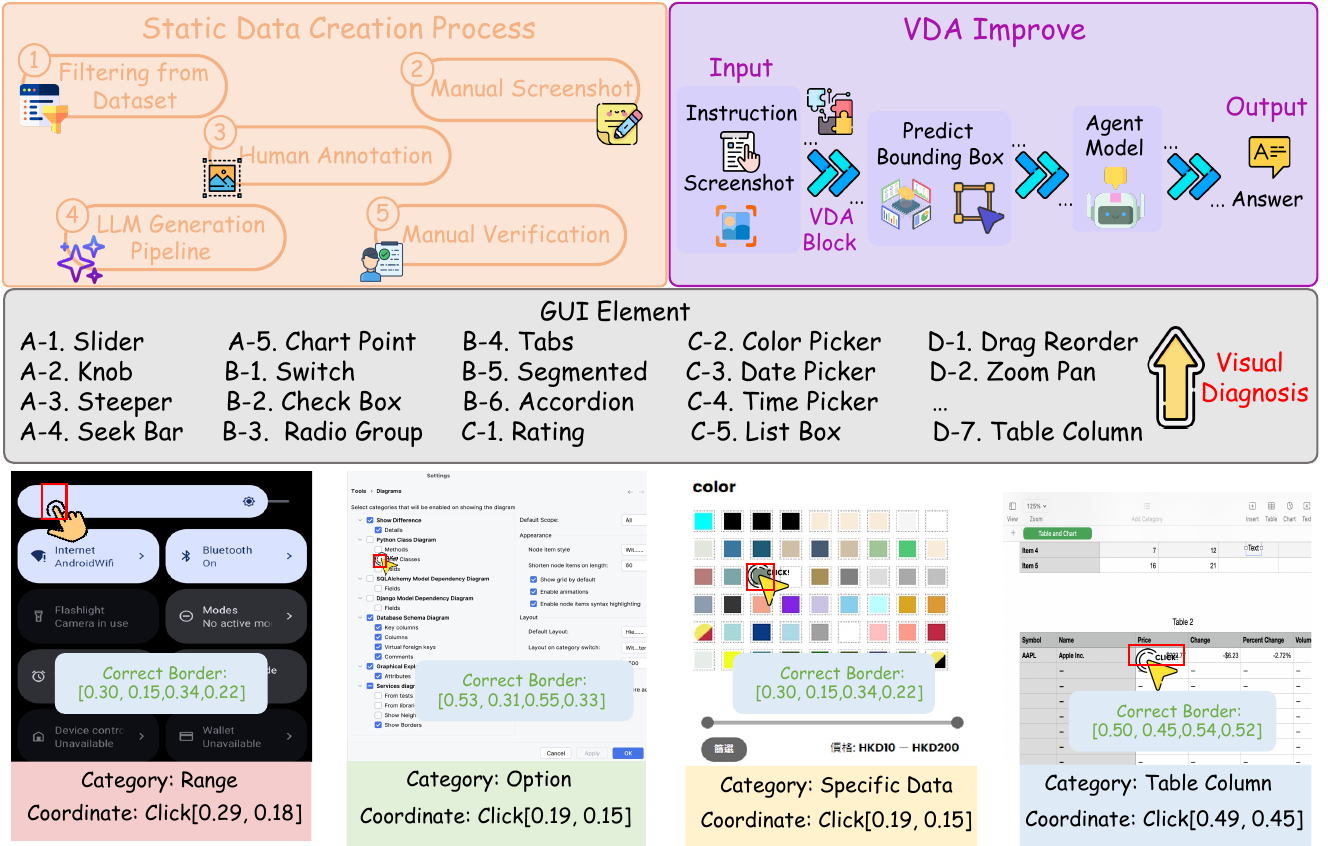}
    \caption{\textbf{Types and Construction Process of the FineState-Bench Dataset, along with an Introduction to the VDA Module:}The figure above shows the data creation process and the VDA assistance mechanism, covering data filtering, annotation, and high-precision bounding box prediction. The middle figure lists the main GUI elements involved in fine-grained control. The bottom figure demonstrates the precise bounding boxes and categories for different types of operations with four examples.}
    \label{fig:FineState-Bench}
\end{figure*}

\section{Related Works}

As our work focuses on the fine-grained, diagnostic evaluation of GUI agents, we briefly overview two closely related research areas: the evolution of GUI agents themselves and the progression of their evaluation methodologies.
 
\subsection{GUI Agents} 
The development of GUI agents, a topic of interest as surveyed by \cite{zhang2024large, nguyen2024guiagents}, is concurrent with the rise of Large Multimodal Models (LMMs) \cite{li2024octopus,wang2024vagi}. Initial studies leveraged general-purpose models like GPT-4V for GUI tasks \cite{yang2023setofmark,apt}, but their localization accuracy and success rates were limited in complex scenarios. To address this, subsequent work shifted to agents designed or fine-tuned for GUI operations, leading to a new generation of specialized agents with evolving design philosophies \cite{qin2023rethinking}. This specialization has manifested across various platforms. For instance, a body of research has focused on mobile agents, with models like Mobile-Agent \cite{wang2024mobile}, MobileFlow \cite{Nong2024MobileFlow}, and AppAgentX \cite{jiang2025appagentxevolvingguiagents} demonstrating increasingly sophisticated capabilities on smartphones. Similarly, agents such as PC-Agent \cite{liu2025pc} and Jedi \cite{Jedi-7B-1080p} have been developed to tackle the complexities of desktop environments. Models such as CogAgent \cite{hong2024cogagent} and ScreenAgent \cite{screenagent} improved visual grounding through pre-training on specialized datasets, while ShowUI \cite{lin2024showui} enhanced generalization with novel UI tokenization and navigation strategies. This evolution towards specialized agents, while enhancing their capabilities, has also exposed the inadequacies of existing evaluation methods, underscoring the urgency of in-depth performance assessment and bottleneck analysis.

\subsection{GUI Agent Evaluation} 
Evaluation methods for GUI agents have also evolved\cite{zheng2024guirobust,zhang2024worldgui,zhao2024screenspotpro, xue2025illusionprogressassessingcurrent, Shlomov2024GroundingPlanning}, shifting focus from success rates to failure diagnostics. Early benchmarks like Mind2Web \cite{deng2023mind2web} relied on static screenshots for offline evaluation. Later, interactive environments like WebArena \cite{zhou2023webarena}, VisualWebArena \cite{koh-etal-2024-visualwebarena}, AITW \cite{gur2024aitw}, AndroidWorld \cite{rawles2024androidworld}, and GAIA \cite{mialon2023gaia} improved realism but suffered from a diagnostic gap, struggling to explain why tasks failed. Classifying GUI defects has been a long-standing challenge. Recently, failure analysis has gained traction, with frameworks like Microsoft's taxonomy for general agents and a fine-grained classification of localization errors \cite{liu2024agentsmith, levy2024st}. This trend spurred a new generation of diagnostic-aware benchmarks: GUI-Robust tests robustness against real-world anomalies; WorldGUI assesses adaptability to varied initial states; ScreenSpot-Pro targets fine-grained localization in professional software; and various testbeds like A-STAR \cite{zhang2024a-star}, Mobile-Bench \cite{deng2024mobile}, and LlamaTouch \cite{zhang2024llamatouch} evaluate agents on general desktop and mobile applications. While these works advance diagnostics, they often target specific failure types. In contrast, our FineState-Bench offers a more universal, structured diagnostic framework. By defining fine-grained failure states, it enables root-cause analysis for any error in any task, yielding more actionable insights for model optimization.

\section{FineState-Bench}

\subsection{Problem Definition}

The core task of FineState-Bench requires a GUI agent to accurately alter the internal state of one or more UI elements to achieve a specific goal based on a natural language instruction $I$. We formally define this task as a sequential decision-making problem: in each task scenario $\tau$, the agent starts from an initial UI state $W_{initial}$ with the objective of executing a sequence of actions $A = \{a_1, a_2, ..., a_T\}$ to reach a final target state $G$. A UI state $W_t$ contains structured information $C_t$ of all interactable controls on the interface, where each control $c \in C_t$ possesses a precise, quantifiable \textbf{fine-grained state value $s(c)$}. The goal $G$ is to drive the state value $s(c^*)$ of a specific target control $c^*$ to the \textbf{target state value $s_{goal}$} required by the instruction.

For complex long-horizon tasks, the agent must maintain awareness of current progress and contextual state throughout task execution, and execute critical operations on the correct UI elements at the correct steps.

\subsection{Dataset Construction and Annotation}

To ensure systematic and comprehensive evaluation, we conducted large-scale screening of modern applications across web, desktop, and mobile platforms, identifying four core interaction categories: (1) Numerical and Range Adjustment (sliders, steppers); (2) State Toggling and Option Selection (switches, checkboxes); (3) Specific Data-Type Selection (date, color pickers); and (4) Content Organization and View Manipulation (drag-and-drop sorting). This taxonomy defines the essential challenges for agents in fine-grained control tasks.

We collected and annotated 2,257 high-quality static samples from the three platforms, each including precise bounding boxes, detailed state information, and natural language instructions. The dataset covers diverse interaction primitives and UI component types, with annotation quality guaranteed through automated pre-filtering and manual verification, providing a solid foundation for fine-grained GUI agent capability evaluation.

For any given component instance $i$, its geometric information includes two key bounding boxes:

\textbf{Locate Bounding Box:} The locate bounding box defines the visible area of the entire UI component. Specifically, $(x_{l1}, y_{l1})$ and $(x_{l2}, y_{l2})$ denote the normalized coordinates of the top-left and bottom-right corners of the component area, with values in the range $[0, 1]$. This bounding box is used to evaluate whether an agent can correctly identify and locate the overall position of the target component.

\begin{equation}
B_{\text{loc}}^{(i)} = [x_{l1}, y_{l1}, x_{l2}, y_{l2}]
\end{equation}

\textbf{Interact Bounding Box:} The interact bounding box specifies the core area within the component that requires precise clicking or operation. The coordinates $(x_{i1}, y_{i1})$ and $(x_{i2}, y_{i2})$ define the range of this actual interactable area, which is typically smaller than or equal to the locate bounding box. This dual-layer design enables separate evaluation of two aspects of the agent's capabilities: \textit{localization ability} (seeing the target) and \textit{precise control ability} (acting accurately).

\begin{equation}
B_{\text{int}}^{(i)} = [x_{i1}, y_{i1}, x_{i2}, y_{i2}]
\end{equation}

\textbf{FineState-Static Benchmark for Static Interactions,} FineState-Static benchmark focuses on static interaction scenarios, requiring agents to precisely alter UI element states through single operations. Its construction process integrates VLM-based pre-filtering with manual curation: initially using a VLM to filter examples with complex interactions from the large-scale os-atlas data source, followed by human auditing of VLM results and manual collection of typical interaction scenarios absent from the source, such as slider dragging and complex date selections. All images are precisely annotated using the LabelImg tool and paired with human-refined natural language instructions. The final dataset consists of high-quality images from web, mobile, and desktop platforms, stored in JSON format as \textit{image + localization + instruction} entries.

\subsection{Multidimensional Evaluation System}

A core advantage of FineState-Bench lies in its multi-dimensional evaluation framework, which quantifies an agent's state perception and fine-grained control on both task completion and execution quality levels. Each task is accompanied by rich metadata, allowing us to diagnose whether failures stem from perception or execution errors.

Our evaluation metrics are designed based on the principle of separation of concerns, decomposing the evaluation process into distinct stages—such as localization and interaction—mirroring the actual workflow of a GUI agent. This hierarchical structure, from basic localization (Loc SR) to precise, single-action state manipulation (SA-Int SR), enables fine-grained diagnosis of agent failures.

This approach provides more scientific and actionable assessment of model capabilities, aligning with GUI automation requirements. Our multi-level metric system serves as both a rigorous evaluation tool and a diagnostic framework, guiding future research toward targeted improvements.

To formally define our evaluation metrics, we first introduce the following notation. For a given test sample $i$, let $p_i$ be the interaction point coordinates predicted by the agent, $B_i$ be the ground-truth bounding box of the target element, $S_{i, \mathrm{pred}}$ be the actual state of the element after interaction, and $S_{i, \mathrm{goal}}$ be the final target state required by the instruction. Let $\mathbb{I}(\cdot)$ be the indicator function, which is 1 if the condition is true, and 0 otherwise.

For static tasks, we design four core metrics to evaluate fundamental localization and interaction capabilities:

\textbf{Locate Success Rate (Loc SR):} Measures whether the agent's predicted interaction point accurately falls within the bounding box of the target UI element. This is the foundation for all successful operations.

\begin{equation}
\mathrm{Loc~SR} = \frac{1}{N} \sum_{i=1}^{N} \mathbb{I}(p_i \in B_i)
\end{equation}

\textbf{Interact Success Rate (Int SR):} Measures whether the agent's interaction successfully brings the state of the target UI element to the intended target state.

\begin{equation}
\mathrm{Int~SR} = \frac{1}{N} \sum_{i=1}^{N} \mathbb{I}(p_i \in B_i \land S_{i, \mathrm{pred}} = S_{i, \mathrm{goal}})
\end{equation}

\textbf{Single-Action Locate Success Rate (SA-Locate SR):} A stricter version of Locate SR, requiring the agent to accurately locate the target in the very first action. Let $p_{i,1}$ be the interaction point of the agent's first action on sample $i$.

\begin{equation}
\mathrm{SA\text{-}Loc~SR} = \frac{1}{N} \sum_{i=1}^{N} \mathbb{I}(p_{i,1} \in B_i)
\end{equation}

\textbf{Single-Action Interact Success Rate (SA-Int SR):} The core metric of FineState-Static. It requires the agent not only to locate accurately but also to successfully drive the UI element to the target state in the first action. Let $S_{i, \mathrm{pred}, 1}$ be the state after the first action's interaction.

\begin{equation}
\tag{6}
\mathrm{SA\text{-}Int~SR} = \frac{1}{N}\sum\nolimits_{i=1}^{N} \mathbb{I}(p_{i,1} \in B_{i} \wedge S_{i,\mathrm{pred},1} = S_{i,\mathrm{goal}})
\end{equation}

This metric serves as our primary benchmark because it captures the complete fine-grained control pipeline: visual understanding, spatial reasoning, and precise motor execution. Unlike isolated metrics that evaluate individual capabilities, SA-Interact SR reflects real-world GUI automation requirements where agents must achieve exact state targets through single, accurate interactions.

FineState-Bench achieves fine-grained evaluation capability through a three-step algorithmic process that enforces mathematical precision:

\begin{algorithm}[H]
\caption{Fine-Grained Evaluation Algorithm}
\begin{algorithmic}[1]
\REQUIRE Task $T_i$ with target state $S_{goal}$, agent prediction $p_{i,1}$, ground truth bounding box $B_i$
\ENSURE Success rate $SR_i \in \{0, 1\}$

\STATE \textbf{Step 1: Spatial Verification}
\IF{$p_{i,1} \notin B_i$}
    \RETURN $SR_i = 0$ \COMMENT{Localization failed}
\ENDIF

\STATE \textbf{Step 2: State Extraction}
\STATE $S_{pred} \leftarrow \text{ExtractState}(T_i, p_{i,1})$ \COMMENT{Get actual state after interaction}

\STATE \textbf{Step 3: Exact State Matching}
\IF{$S_{pred} = S_{goal}$}
    \RETURN $SR_i = 1$ \COMMENT{Precise state achieved}
\ELSE
    \RETURN $SR_i = 0$ \COMMENT{State mismatch, no approximation allowed}
\ENDIF

\end{algorithmic}
\end{algorithm}

\section{Benchmark Analysis}

\subsection{Statistical Analysis}

\begin{wrapfigure}{r}{0.49\textwidth} 
    \centering
    \vspace{-15pt} % 可选：用于微调图片与上方文本的垂直距离，根据需要调整或删除
    \begin{minipage}[t]{0.49\linewidth} % 子图宽度，相对于 wrapfigure 的宽度
        \centering
        \includegraphics[width=\textwidth]{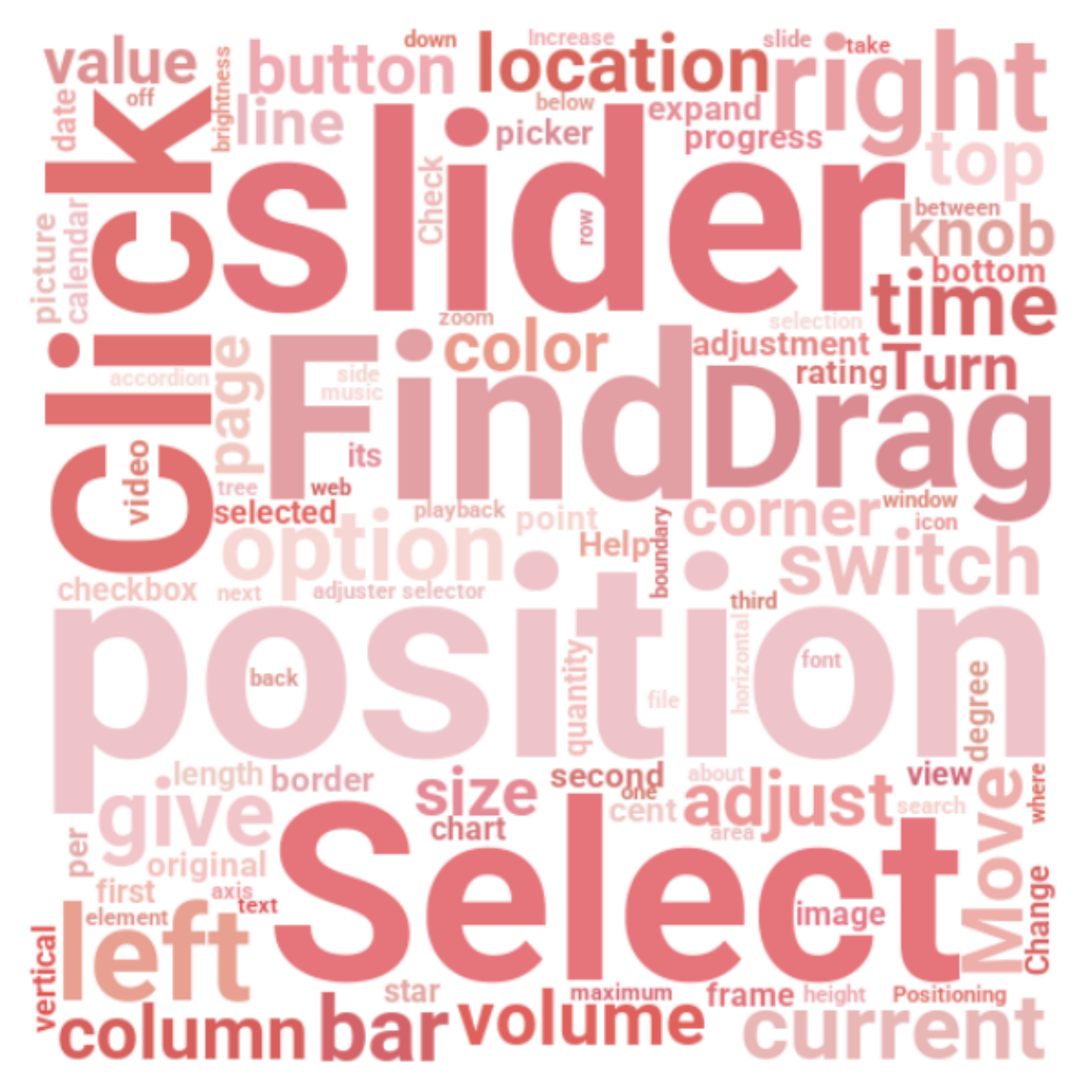} % 图片路径
        \caption*{\footnotesize(a) Instruction Word Cloud}
    \end{minipage}
    \hfill % 在两个子图之间提供弹性空间
    \begin{minipage}[t]{0.49\linewidth} % 子图宽度
        \centering
        \includegraphics[width=\textwidth]{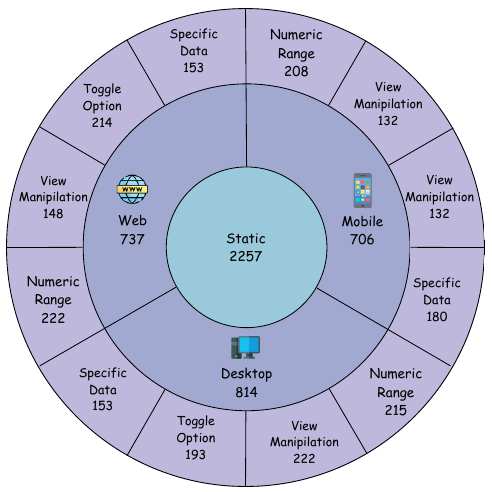} % 图片路径
        \caption*{\footnotesize (b) Task Type Distribution}
    \end{minipage}
    \caption{Visualization of instruction keywords and task type composition in FineState-Bench.}
    \label{fig:wordcloud_pie} % 更新后的label
\end{wrapfigure}

The composition and platform distribution of FineState\textendash Bench are illustrated in Figure~3(b).  
This benchmark includes a total of 2,257 static tasks, distributed across three major platforms: desktop (814), web (737), and mobile (706), covering four major categories of interactive components such as numerical range adjustment, option switching, specific data operations, and view operations, aiming to systematically measure the fundamental fine control ability of the model.

% \begin{figure}[h!]
%     \centering
%     \begin{minipage}[t]{0.45\textwidth}
%         \centering
%         \includegraphics[width=\textwidth]{figure/WordCloud.pdf}
%         \caption*{(a) Instruction Word Cloud}
%     \end{minipage}
%     \hfill
%     \begin{minipage}[t]{0.44\textwidth}
%         \centering
%         \includegraphics[width=\textwidth]{figure/Piechart.pdf}
%         \caption*{(b) Task Type Distribution}
%     \end{minipage}
%     \caption{Visualization of instruction keywords and task type composition in FineState-Bench.}
%     \label{fig:wordcloud_pie}
% \end{figure}

Additionally, to showcase the richness and challenge of the instructions, we visualized all task commands.  
As shown in Figure~3(a), the word cloud vividly displays high-frequency core verbs (e.g., slider, select, click, drag, adjust) and key entities (volume, brightness, location, button, time, color).  
This visualization not only highlights the benchmark's central focus on fine-grained manipulation, but also reveals the diversity and practicality of its instructions in simulating real-world tasks.

\begin{table*}[!t]
  \centering
  \caption{Comparison of FineState-Bench with existing GUI agent benchmarks, analyzing them across key dimensions such as platform coverage, instruction type, task granularity, and dataset size. FineState-Bench is distinguished by its cross-platform nature, its focus on fine-grained interactions, and its ability to differentiate between perception and reasoning failures. In this context, \textit{Que.} stands for Question, and \textit{Gra.} stands for Task Granularity type indicates the type of distinction for the dataset. Types refers to the number of different categories of dataset components, websites, or apps.}
  \label{tab:dataset_comparison}
  
  \newcommand{\gcheck}{\ding{51}}

  \resizebox{\textwidth}{!}{%
  \begin{tabular}{c ccccc c c c}
    \toprule
     \multirow{2}{*}{\textbf{Benchmark}} & \multicolumn{3}{c}{\textbf{Environment (Env.)}}& \multicolumn{3}{c}{\textbf{Instruction (Ins.)}} & \multirow{2}{*}{\textbf{Types}} & \multirow{2}{*}{\textbf{Number}} \\
     & Desktop& Mobile& Website& Que. & Gra. & Ins. Level & & \\
    \midrule
     Mind2Web~\cite{deng2023mind2web} 
    & & & \gcheck &  & \gcheck & High & 31 & 2350 \\
     WebArena~\cite{zhou2023webarena} 
    & & & \gcheck & \gcheck & & High & 4 & 812 \\
     AndroidWorld~\cite{rawles2024androidworld} 
    & & \gcheck & & & \gcheck & Mid & 20 & 116 \\
     VisualWebArena~\cite{koh-etal-2024-visualwebarena} 
    & & &\gcheck  &  & & High & 3 & 910 \\
     GAIA~\cite{mialon2023gaia} 
    & \gcheck & &  & \gcheck &  & High & 5 & 279 \\
     WebVoyager~\cite{he-etal-2024-webvoyager} 
    & & & \gcheck & & \gcheck & Low/Mid & 15 & 643 \\
     OSWORLD-G~\cite{OSWORLD-G} 
    & \gcheck & & & \gcheck & \gcheck & High & 32  & 564 \\
    \midrule
     FineState-Bench (Ours) 
    & \gcheck & \gcheck & \gcheck & \gcheck & \gcheck & High & 23 & 2257 \\
    \bottomrule
  \end{tabular}%
  }
\end{table*}

\subsection{Comparative Analysis with Existing Benchmarks}

To highlight the unique contributions of FineState\textendash Bench in the field of GUI-agent evaluation, we conduct a systematic, multi-dimensional comparison with mainstream benchmarks.  
As shown in Table~1, we examine platform coverage, environment realism, task and evaluation granularity, and—crucially—diagnostic capability, providing an in-depth analysis of FineState-Bench alongside representative works such as Mind2Web, WebArena, AndroidWorld, VisualWebArena, GAIA, and WebVoyager.

Existing benchmarks have common limitations: most are confined to a single platform and use a coarse, outcome-oriented evaluation, making it difficult to investigate the root causes of task failures and creating an evaluation gap. Even advanced dynamic environments offer only indirect diagnostic capabilities, lacking the ability to perform fine-grained, attributable analysis of failures. OSWORLD-G covers 564 samples and 32 types of UI elements, and includes fine-grained manipulation as one of its five major capability dimensions, but its evaluation method still belongs to an integrated capability test and does not delve into a deep understanding of fine-grained manipulation.

In contrast, FineState-Bench provides a solution through its unique design. First, it spans the Web, Mobile, and Desktop platforms, shifting the evaluation focus from high-level task success to the execution quality of interaction primitives. Second, it establishes a multi-dimensional evaluation system covering success, efficiency, and precision. Its core contribution is a VDA that systematically distinguishes between perception failures and reasoning failures, shifting the research paradigm from "what a model can do" to a root-cause analysis of "why a model fails".

\section{VDA: Diagnosing Vision Bottlenecks}

To investigate GUI agent task failure, whether due to interaction inference defects or insufficient visual perception, we designed a diagnostic experiment using the control variable method with the plug-and-play VDA module, providing target localization information. We compared the standard agent with the visual enhanced version equipped with VDA. Since the core inference logic is consistent, the performance difference stems from improved visual ability. This method lets us pinpoint the agent's visual bottleneck.

To delineate whether GUI agent failures stem from deficient interaction reasoning or insufficient visual perception, we designed a diagnostic experiment based on the control variable methodology. At the core of this experiment is a VDA module, providing the agent with target localization information. The experiment compares a standard agent against a vision-enhanced counterpart augmented by the VDA. Since the core reasoning logic remains identical, the performance disparity can be attributed to the enhancement of visual capabilities. This approach enables us to dissect and pinpoint the agent's visual bottlenecks.

\begin{table*}[!h]
\centering
\caption{Baseline evaluation on `FineState-Static`. We report four metrics: Locate SR / Interact SR / SA-Locate SR / \textbf{SA-Interact SR} (\%). SA-Interact SR (shown in bold) represents our primary evaluation criterion, requiring simultaneous accurate localization and precise state manipulation in a single action - the gold standard for fine-grained GUI control assessment.}
\label{tab:static_results_en_corrected}
\resizebox{\textwidth}{!}{
\begin{tabular}{lcccccc}
\toprule

\multicolumn{2}{c}{\textbf{Model Category}} & \textbf{Model Name} & \textbf{Mobile} & \textbf{Web} & \textbf{Desktop} & \textbf{AVG} \\
\midrule

\multirow{9}{*}{Generalist Models} &

\multirow{3}{*}{Closed-source} & GPT-4o & 31.0/6.2/9.1/\textbf{6.2} & 22.8/4.5/7.0/\textbf{4.5} & 20.6/2.2/4.4/\textbf{2.2} & 24.8/4.3/6.8/\textbf{4.3} \\

& & Claude-3.5-Sonnet & 31.7/11.5/13.7/\textbf{11.5} & 15.2/1.9/4.6/\textbf{1.9} & 22.0/3.4/5.4/\textbf{3.4} & 23.0/5.6/7.9/\textbf{5.6} \\
& & Gemini-2.5-Flash & 49.4/17.6/21.1/\textbf{17.6} & 47.0/11.8/16.8/\textbf{11.8} & 12.7/0.7/3.8/\textbf{0.7} & 36.4/10.0/13.9/\textbf{10.0} \\
\cmidrule(l){2-7} 
& \multirow{5}{*}{Open-source} & OS-Atlas-7B\cite{OS-Atlas} & 47.5/12.8/18.7/\textbf{12.8} & 33.2/7.5/9.2/\textbf{7.5} & 45.3/9.8/15.2/\textbf{9.8} & 42.0/10.0/14.4/\textbf{10.0} \\
& & CogAgent-9B\cite{hong2024cogagent} & 17.7/1.8/2.5/\textbf{1.8} & 24.1/6.4/13.7/\textbf{6.4} & 29.4/2.3/3.5/\textbf{1.2} & 23.7/3.5/6.6/\textbf{3.1} \\
& & UGround-7B\cite{gou2024uground} & 50.7/19.6/22.4/\textbf{19.6} & 62.0/32.8/62.0/\textbf{32.8} & 46.3/16.0/25.4/\textbf{16.0} & 53.0/22.8/36.6/\textbf{22.8} \\
& & Jedi-7B-1080p\cite{Jedi-7B-1080p} & 13.3/1.6/3.1/\textbf{1.5} & 12.7/8.3/12.7/\textbf{8.3} & 12.2/0.8/1.5/\textbf{0.8} & 12.7/3.6/5.8/\textbf{3.5} \\
& & ShowUI-2B\cite{lin2024showui} & 20.3/5.2/6.7/\textbf{5.2} & 26.7/5.3/26.7/\textbf{5.3} & 30.3/3.2/9.1/\textbf{3.2} & 25.8/4.6/14.2/\textbf{4.6} \\
\midrule

\multirow{5}{*}{Specialist Models} &
\multirow{2}{*}{Mobile} & MobileVLM V2-3B & 16.5/1.6/2.5/\textbf{1.6} & 16.1/2.6/4.0/\textbf{2.6} & 22.9/1.9/4.3/\textbf{1.9} & 18.5/2.0/3.6/\textbf{2.0} \\
& & MobileVLM V2-7B & 18.5/1.9/2.9/\textbf{1.9} & 26.7/2.6/5.9/\textbf{2.6} & 26.3/3.7/5.2/\textbf{3.7} & 23.8/2.7/4.7/\textbf{2.7} \\
\cmidrule(l){2-7}
& Web & Holo1-7B\cite{Holo1} & 3.5/1.6/2.6/\textbf{1.6} & 12.2/1.3/3.8/\textbf{1.3} & 7.1/3.6/4.0/\textbf{2.6} & 7.6/2.2/3.5/\textbf{1.8} \\
\cmidrule(l){2-7}
& Desktop & AgentCPM-GUI-8B\cite{AgentCPM-GUI} & 42.4/18.2/32.4/\textbf{18.2} & 30.6/7.9/30.6/\textbf{7.9} & 34.6/7.1/9.3/\textbf{7.1} & 35.9/11.1/24.1/\textbf{11.1} \\
\bottomrule
\end{tabular}%
}
\end{table*}

As an external visual localization system, VDA uses a \textit{describe-then-locate} two-stage process. First, In the first phase, it finds the target UI element based on the instruction and screenshot, and describes the properties of the target UI element, the internal situation of the UI element, and its position relative to the image. Second, it integrates all information to predict a high-precision bounding box. This chain-of-thought process enhances visual localization robustness through language.

\textbf{VDA Implementation Details:} The VDA operates through a structured two-stage pipeline: 

\textbf{Stage 1 - Contextual Description:} Given a screenshot and natural language instruction, GPT-4o generates a detailed textual description of the target UI element, including: (i) functional purpose and current state, (ii) visual characteristics and spatial relationships, (iii) distinguishing features from similar elements. This produces rich semantic context that serves as a visual anchor.

\textbf{Stage 2 - Precision Localization:} Integrating the instruction, screenshot, and Stage 1 description, GPT-4o predicts high-precision bounding boxes in normalized coordinates [x1, y1, x2, y2]. These coordinates are then provided to the target model as additional input, simulating ideal visual grounding capabilities.

\textbf{Integration Mechanism:} VDA operates as a plug-and-play preprocessor - target models receive their standard inputs (screenshot + instruction) plus the VDA-generated coordinates, enabling direct performance comparison with baseline conditions while maintaining identical reasoning pathways.

To systematically evaluate existing GUI agents on fine-grained tasks and diagnose their potential performance bottlenecks, a goal shared by other recent benchmarking efforts \cite{bonatti2024arena, chen2025spabench}, we designed three interconnected experiments. This section will follow a four-part structure: Experimental Setup - Baseline Evaluation - Dynamic Evaluation - Diagnostic Study, detailing the experimental process, results, and analysis.

\section{Experiments and Analysis}

\subsection{Experimental Setup}

To assess the fine-grained control capabilities of contemporary GUI agents, we conduct a comprehensive evaluation of 13 representative models, encompassing both commercial closed-source and open-source research systems. Our selection covers three categories: (1) closed-source commercial models, including GPT-4o\cite{gpt4o}, Claude-3.5-Sonnet\cite{claude_3_5_sonnet}, and Gemini-2.5-Flash\cite{gemini_2_5_flash}; (2) specialized open-source GUI agents, such as UGround~\cite{gou2024uground}, OS-Atlas~\cite{OS-Atlas}, and CogAgent~\cite{hong2024cogagent}; and (3) platform-specific models, including the MobileVLM series~\cite{wu2024mobilevlm} for mobile scenarios, Holo1-7B for web environments, and AgentCPM-GUI for desktop applications. This stratified selection enables a thorough analysis across different capability levels and specialization domains, providing an overview of the current state of fine-grained GUI control on desktop, web, and mobile platforms. By evaluating these models, we aim to elucidate the strengths and limitations of state-of-the-art GUI agents in performing precise and complex user interface operations.

\subsection{Baseline Deficiencies in Fine-Grained Control}

In order to accurately assess the real capabilities of the most advanced GUI agents in performing basic and refined operations, we first conducted a comprehensive baseline evaluation on the FineState-Static benchmark.As shown in Table 2, the results reveal a common limitation among contemporary advanced models: all tested models, whether closed-source general-purpose models or specialized platform-specific ones, underperformed on our core metric, SA-Interact SR (Single-Action Interaction Success Rate). This finding strongly indicates that current GUI agents generally lack stable and precise fine-grained control capabilities, even when handling the seemingly simple static scenarios that constitute the basis for complex tasks.

\subsection{Cross-Platform Performance Comparison}

Cross-platform analysis reveals striking performance disparities that challenge conventional assumptions about GUI complexity. Contrary to intuitive expectations, performance varies significantly across platforms, with some models showing better results on mobile interfaces while others excelt on web or desktop environments. For instance, UGround-7B achieves 32.8\% success on web, 19.6\% on mobile, and 16.0\% on desktop, while Gemini-2.5-Flash shows the opposite pattern with 17.6\% on mobile versus 11.8\% on web, highlighting platform-specific optimization opportunities and the need for platform-aware model development.

\section{Diagnostic Study: Pinpointing the Root Cause of Performance Bottlenecks}

The results of the first two experiments reveal the widespread failure of existing GUI agents in fine-grained control tasks (Table 2) and their performance degradation in dynamic scenarios (Table 3), thus systematically confirming the existence of an evaluation gap. To bridge this diagnostic gap, we designed a third experiment to investigate the root causes of task failure.

\subsection{Quantitative Evidence of the Vision Bottleneck}

\textbf{Methodological Clarification:} VDA is designed as a diagnostic tool, not a performance enhancement system. By providing models with idealized localization information through GPT-4o, we create controlled experimental conditions that isolate visual grounding effects from other cognitive processes. The performance improvements observed represent the quantifiable impact of visual limitations on overall task success, establishing causal evidence for our core hypothesis.

To assess the impact of high-fidelity visual grounding on multimodal model performance, we employed our VDA framework, which leverages GPT-4o to provide precise localization guidance. This setup enables controlled experiments that isolate the effect of visual grounding from other cognitive processes. We conducted comparative evaluations on three representative models: Gemini-2.5-Flash, ShowUI-2B, and OS-Atlas-7B.

As shown in Table~4, VDA intervention yields dramatic performance stratification: Gemini-2.5-Flash demonstrates substantial improvements (14.9\% average gain), while ShowUI-2B shows minimal enhancement (3.7\% gain). This disparity reveals a critical insight—visual grounding improvements are ultimately bounded by models' intrinsic reasoning and execution capabilities. Superior visual input cannot compensate for fundamental deficiencies in logical processing or motor control precision.

\begin{table}[h]
\centering
\caption{Overall impact of the VDA on `SA-Interact SR` (\%). `Gain \(\Delta\)` quantifies the performance improVDAent from enhanced visual capabilities.}
\label{tab:VDA_overall_en}
\resizebox{\columnwidth}{!}{%
\begin{tabular}{cccc}
\toprule
\textbf{Model} & \textbf{Mobile} & \textbf{Web} & \textbf{Desktop}  \\
\midrule
Gemini-2.5-Flash & 17.6 & 11.8 & 0.7  \\
ShowUI-2B     & 5.2 & 5.3 & 3.2  \\
OS-Atlas-7B    & 12.8 & 7.5 & 9.8 \\
\midrule
Gemini-2.5-Flash(VDA-Gemini-2.5-Flash) & 29.8 & 27.2 & 15.4  \\
ShowUI-2B(VDA-Gemini-2.5-Flash)     & 5.8 & 12.3 & 7.5  \\
OS-Atlas-7B(VDA-Gemini-2.5-Flash)    & 18.9 & 18.2 & 19.1 \\
\midrule
Gemini-2.5-Flash(VDA-GPT4o) & 29.8 & 26.6 & 20.9  \\
ShowUI-2B(VDA-GPT4o)     & 7.3 & 7.2 & 9.5  \\
OS-Atlas-7B(VDA-GPT4o)    & 15.9 & 14.2 & 13.1 \\

\bottomrule
\end{tabular}%
}
\end{table}

\subsection{Ablation Study of VDA}

To validate the effectiveness of the two-stage \textit{describe-and-locate} design in our VDA framework, we conducted an ablation study (see Table~4) comparing two configurations: using only the coordinate prediction stage, and employing the full pipeline with both description and localization.

The results show that coordinate prediction alone yields a substantial performance boost, confirming that precise localization is key to VDA's success. However, the best results are achieved when both stages are combined, as the initial description provides essential context for accurate localization. This demonstrates that the two-stage process—first describing, then locating—offers complementary benefits and is critical for optimal fine-grained control.

\begin{table}[h]
\centering
\caption{Ablation study of VDA-Flash's internal components on Gemini 2.5 Flash. We measure the contribution of each stage to the final `SA-Interact SR` (\%).}
\label{tab:VDA_ablation_en}
\resizebox{\columnwidth}{!}{%
\begin{tabular}{cccc}
\toprule
\textbf{VDA Configuration} & \textbf{Mobile} & \textbf{Web} & \textbf{Desktop}\\
\midrule
Baseline (No VDA) & 17.6 & 11.8 & 0.7 \\
 Stage Two Only (Description)& 17.9 & 11.9 & 0.8 \\
Stage Two Only (Coordinate Prediction) & 26.1 & 24.7 & 13.1 \\
Full Two-Stage (Description + Coordinate) & 29.8 & 27.2 & 15.4 \\
\bottomrule
\end{tabular}%
}
\end{table}

To complement the quantitative results, we conducted a qualitative analysis of failure modes and their VDA remediation effectiveness. Our analysis identifies four primary failure categories: Localization Ambiguity (85\% error reduction with VDA), Visual Feature Confusion (72\%), Fine-Grained State Perception Failure (45\%), and Interaction Context Ignorance (23\%). This hierarchy reveals that while spatial reasoning deficits are largely addressable via enhanced visual grounding, higher-order contextual understanding remains a persistent architectural challenge requiring innovations beyond improved visual input processing. Detailed case studies and failure mode analysis are provided in the Supplementary materials.

\section{Conclusion}

Our work identifies a fundamental limitation in current GUI agent evaluation: mainstream benchmarks rely on coarse-grained criteria, overlooking the need for fine-grained state awareness and precise control, which results in an overestimation of model capabilities. To bridge this \textit{evaluation gap}, we present FineState-Bench, the first open-source benchmark and diagnostic framework designed for comprehensive, fine-grained assessment across desktop, web, and mobile platforms. FineState-Bench introduces multi-level evaluation metrics and a dual-bounding-box annotation scheme, enabling detailed analysis of both localization and control abilities. Furthermore, we propose a plug-and-play VDA that facilitates quantitative diagnosis of visual grounding bottlenecks. Our findings reveal that the primary challenge for current GUI agents lies in low-level visual localization, rather than high-level planning. By shifting the focus from task completion to the underlying causes of failure, FineState-Bench establishes a new standard for realistic and actionable evaluation and highlights the importance of foundational visual capabilities for advancing next-generation GUI agents.

\section{Acknowledgement}

Thanks to Xie Yu, Yuheng Song, Jinghang Zhong, Yuxin Ye, and Mengran Liu for their help with annotating the data for our project.

\section*{References}

\bibliographystyle{plainnat}
\bibliography{neurips_2024}

\appendix

\section{Appendix}

\subsection{A.1 Dataset Architecture and Content Showcase}

This section presents a comprehensive overview of the FineState-Bench dataset architecture and showcases representative examples from our multi-platform, multi-task dataset collection.

\subsubsection{A.1.1 Dataset Overall Architecture}

The FineState-Bench dataset adopts a multi-platform, multi-task, multi-component hierarchical structure design:

\textbf{Platform Dimension:} The dataset spans three major platforms - Web, Mobile, and Desktop - ensuring comprehensive coverage of modern GUI environments.

\textbf{Task Dimension:} We organize the dataset into four main task categories:
\begin{enumerate}
    \item \textbf{Numeric Range Adjustment (A):} Controls that manipulate continuous numerical values including Sliders, Knobs, Steppers, Seek Bars, and Chart Points. These components require precise value manipulation and state awareness.
    
    \item \textbf{Toggle Option Selection (B):} Binary or multi-option selection controls including Switches, Checkboxes, Radio Groups, Tabs, Segmented Controls, and Accordions. These components focus on discrete state transitions.
    
    \item \textbf{Specific Data Selection (C):} Specialized input controls for specific data types including Rating components, Color Pickers, Date Pickers, Time Pickers, and Listbox/Dropdown menus. These require domain-specific understanding and precise interaction.
    
    \item \textbf{View Manipulation (D):} Controls that modify the visual layout or organization including Drag \& Reorder, Zoom \& Pan, Resizable Panes, Carousels, Tree Views, Splitters, and Table Columns. These components require spatial reasoning and complex interaction sequences.
\end{enumerate}

\subsubsection{A.1.2 Data Scale Statistics}

Based on our comprehensive analysis of the English JSON files across all platforms, the FineState-Bench dataset contains:

\textbf{Overall Data Volume:}
\begin{itemize}
    \item Total Samples: 2,257 high-quality annotated instances
    \item Platform Coverage: 3 major platforms with balanced distribution
    \item Task Categories: 4 main categories with 23 distinct component types
\end{itemize}

\textbf{Platform-wise Distribution:}
\begin{itemize}
    \item Desktop: 814 samples (36.1\%)
    \item Web: 737 samples (32.7\%)
    \item Mobile: 706 samples (31.3\%)
\end{itemize}

\textbf{Task Category Distribution by Platform:}

\begin{table}[h]
\centering
\caption{Platform × Task Category Distribution}
\resizebox{\textwidth}{!}{%
\begin{tabular}{lcccc}
\toprule
\textbf{Category} & \textbf{Desktop} & \textbf{Mobile} & \textbf{Web} & \textbf{Total} \\
\midrule
Numeric Range Adjustment & 219 & 213 & 221 & 653 (28.9\%) \\
Toggle Option Selection & 198 & 191 & 186 & 575 (25.5\%) \\
Specific Data Selection & 197 & 130 & 155 & 482 (21.4\%) \\
View Manipulation & 200 & 172 & 175 & 547 (24.2\%) \\
\midrule
\textbf{Total} & \textbf{814} & \textbf{706} & \textbf{737} & \textbf{2,257} \\
\bottomrule
\end{tabular}%
}
\end{table}

\subsubsection{A.1.3 Representative Data Content Examples}

To demonstrate the diversity and complexity of our dataset, we present representative examples from each major task category. Each example includes the original screenshot, component metadata, natural language instructions, and precise state transition requirements.

\textbf{Numeric Range Adjustment Examples:}

\begin{figure}[htbp!]
    \centering
    \begin{minipage}{0.3\textwidth}
        \centering
        \includegraphics[width=\textwidth]{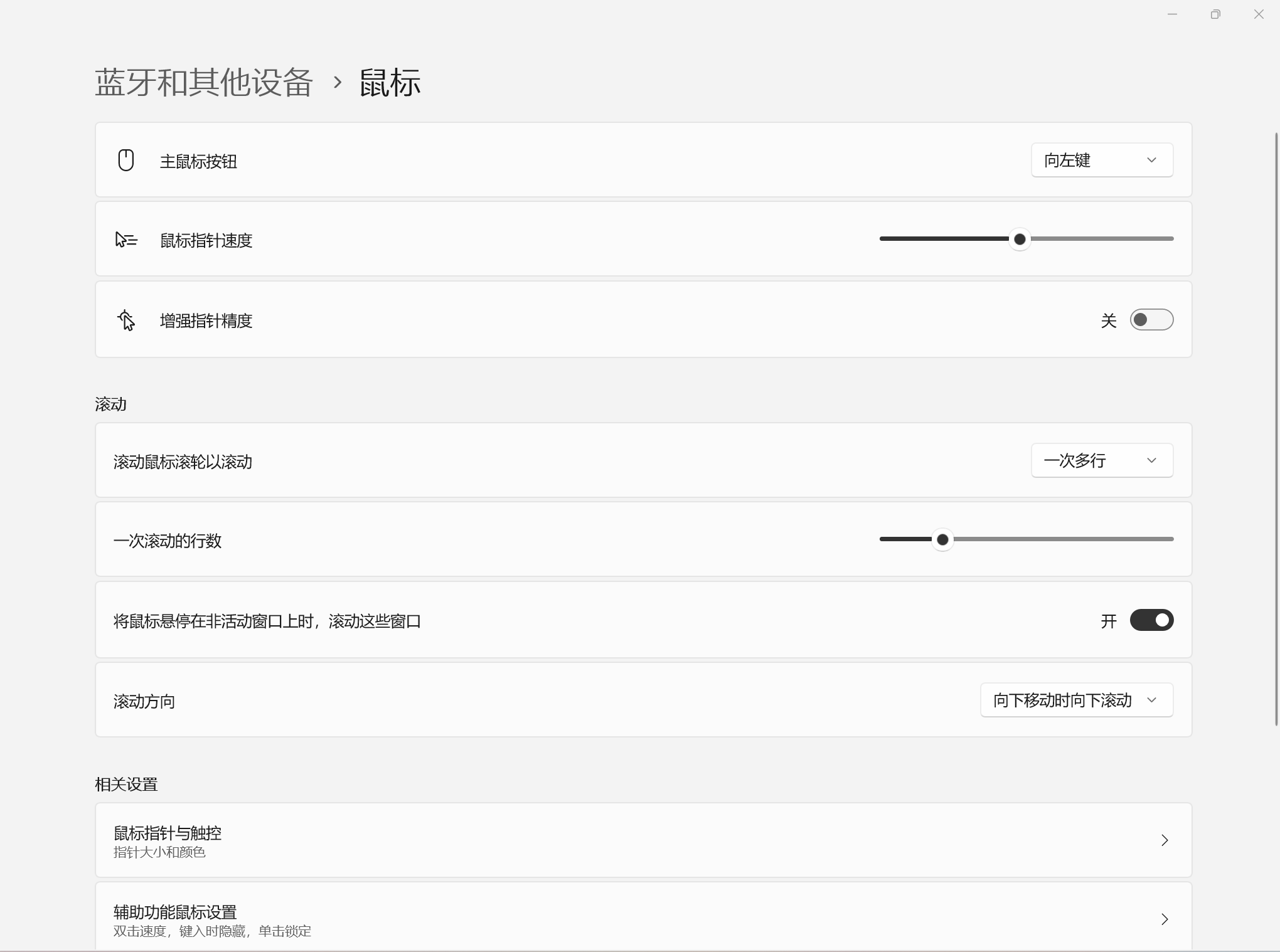}
        \caption*{Slider (Desktop)}
    \end{minipage}
    \hfill
    \begin{minipage}{0.3\textwidth}
        \centering
        \includegraphics[width=\textwidth]{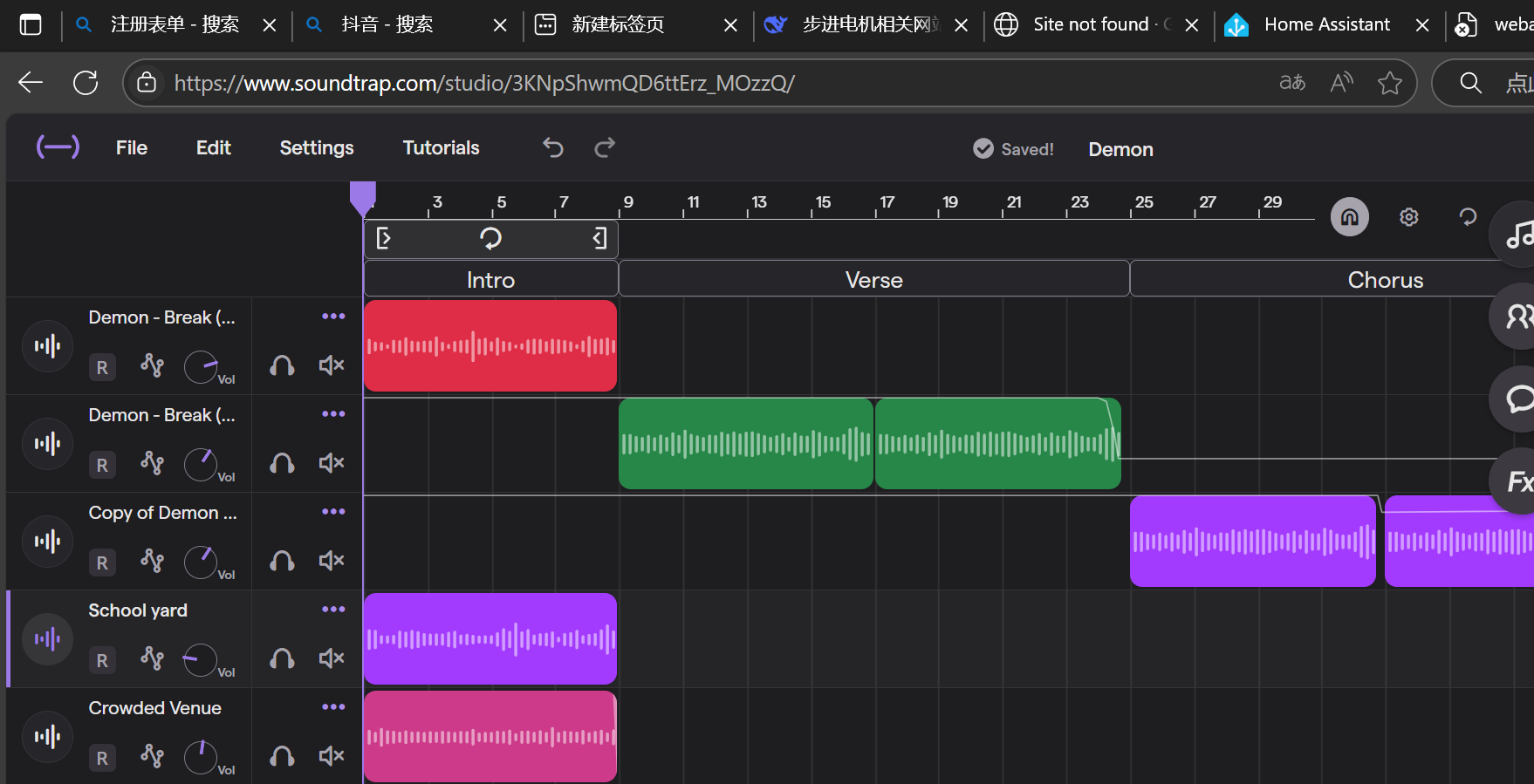}
        \caption*{Knob (Web)}
    \end{minipage}
    \hfill
    \begin{minipage}{0.3\textwidth}
        \centering
        \includegraphics[width=\textwidth]{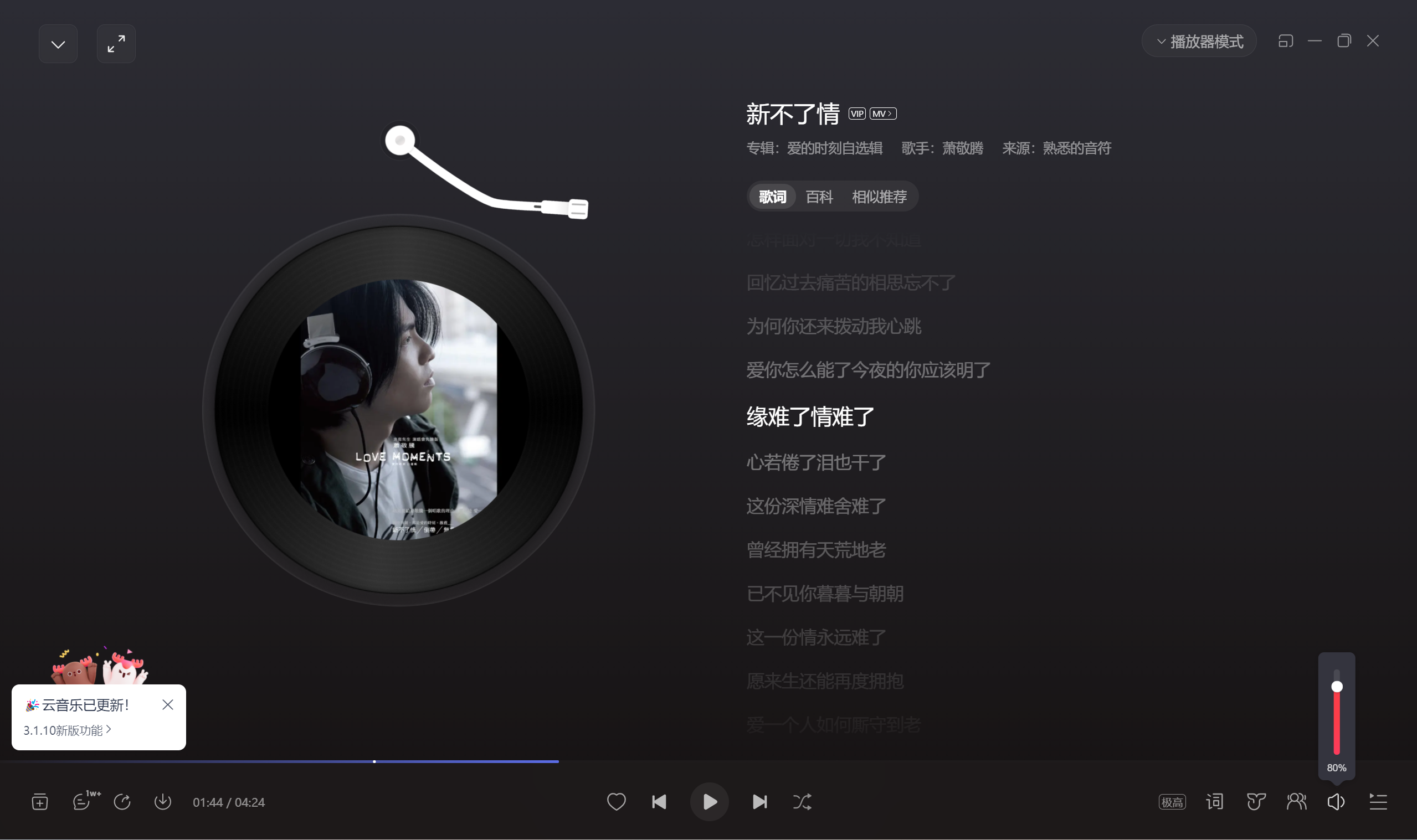}
        \caption*{Seek Bar (Web)}
    \end{minipage}
\end{figure}

\textbf{Toggle Option Selection Examples:}

\begin{figure}[htbp!]
    \centering
    \begin{minipage}{0.3\textwidth}
        \centering
        \includegraphics[width=\textwidth]{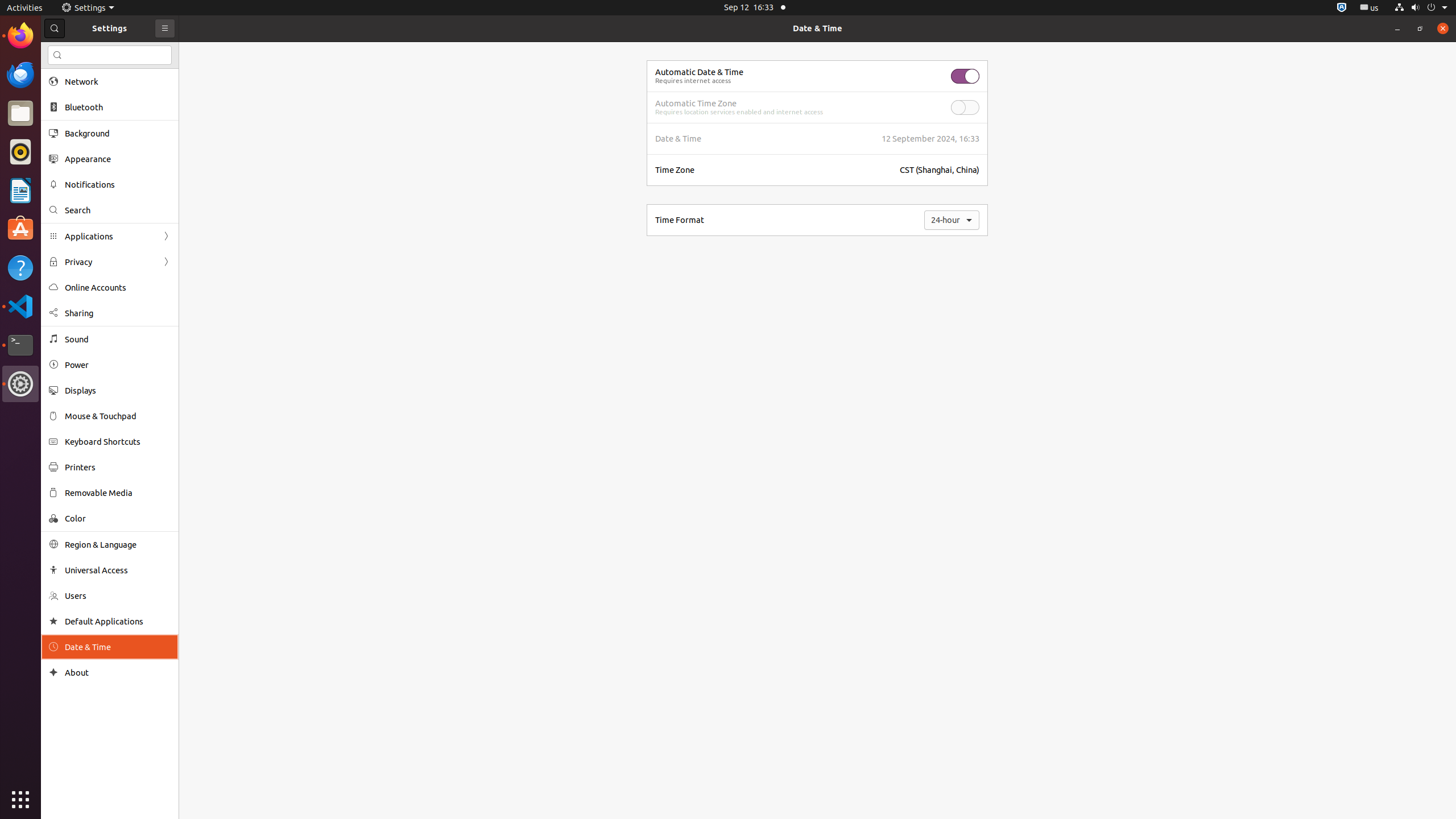}
        \caption*{Switch (Desktop)}
    \end{minipage}
    \hfill
    \begin{minipage}{0.3\textwidth}
        \centering
        \includegraphics[width=\textwidth]{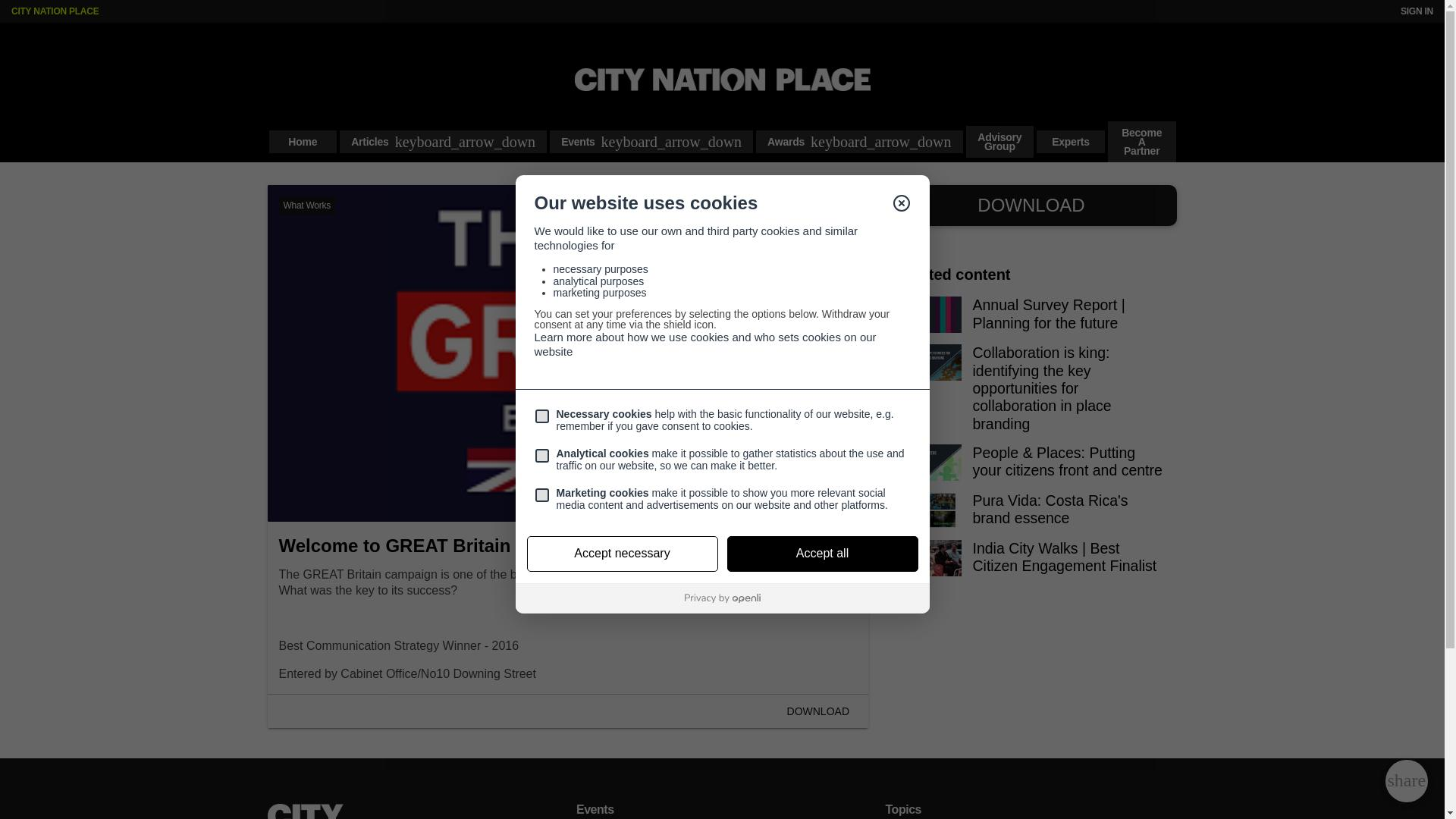}
        \caption*{Checkbox (Web)}
    \end{minipage}
    \hfill
    \begin{minipage}{0.3\textwidth}
        \centering
        \includegraphics[width=\textwidth]{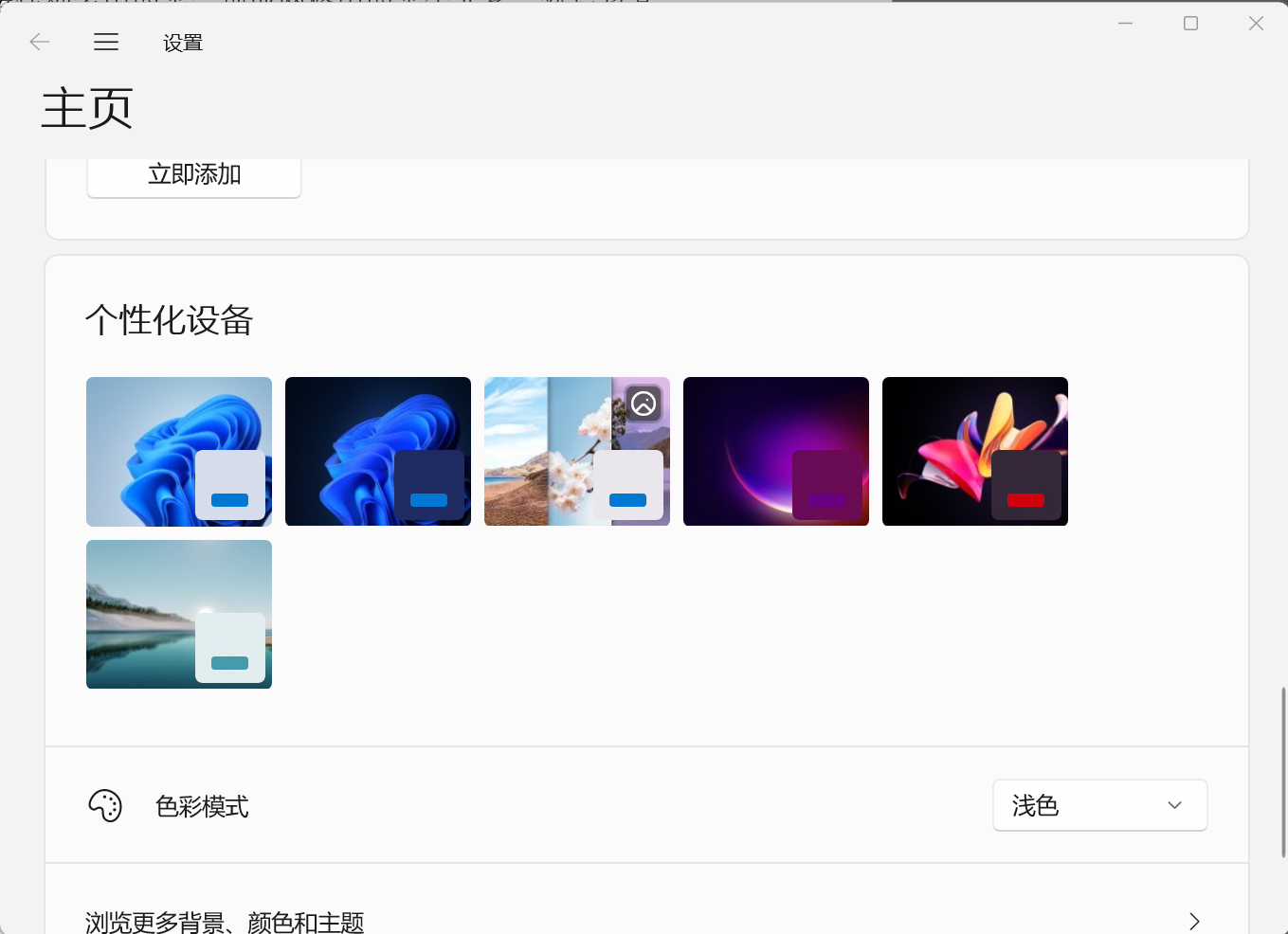}
        \caption*{Radio Group (Desktop)}
    \end{minipage}
\end{figure}

\textbf{Specific Data Selection Examples:}

\begin{figure}[htbp!]
    \centering
    \begin{minipage}{0.45\textwidth}
        \centering
        \includegraphics[width=\textwidth]{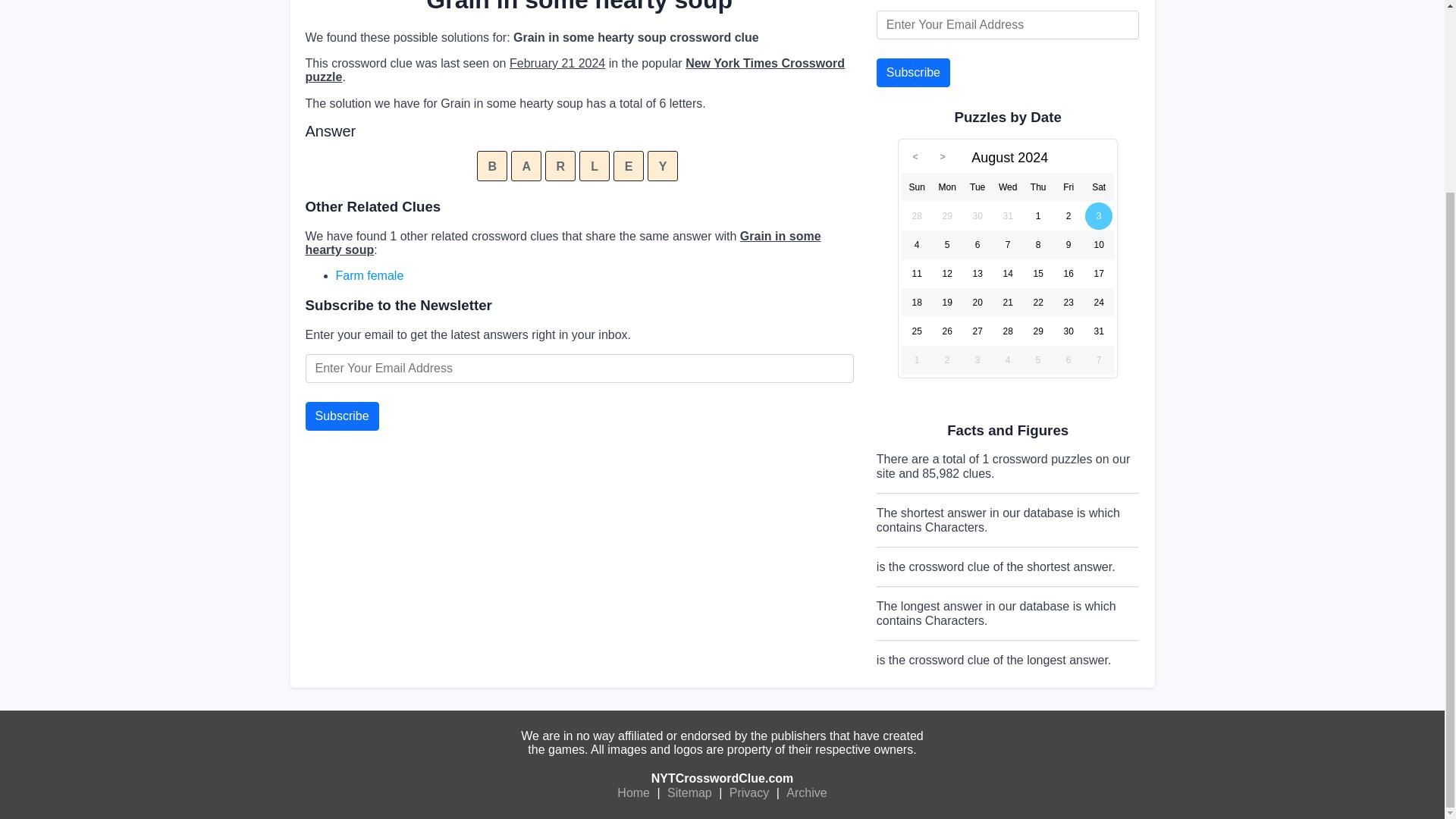}
        \caption*{Date Picker (Web)}
    \end{minipage}
    \hfill
    \begin{minipage}{0.45\textwidth}
        \centering
        \includegraphics[width=\textwidth]{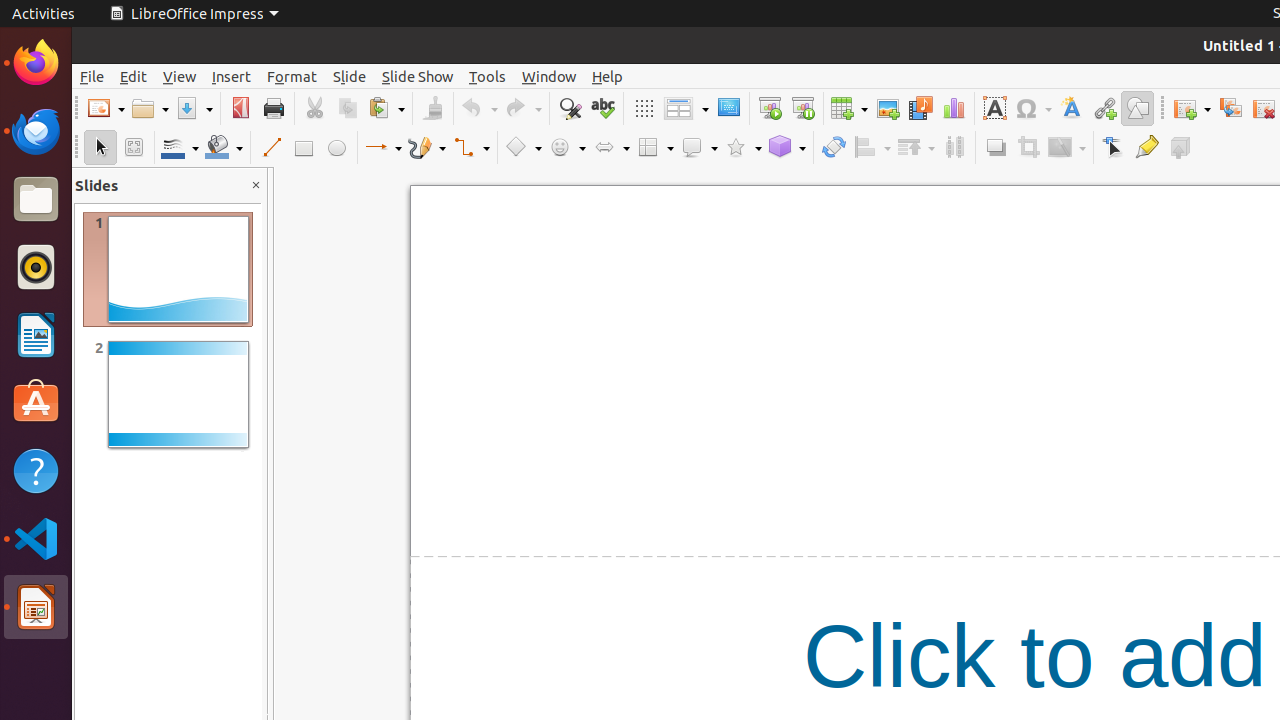}
        \caption*{Drag \& Reorder (Desktop)}
    \end{minipage}
\end{figure}

\textbf{View Manipulation Examples:}

\begin{figure}[htbp!]
    \centering
    \begin{minipage}{0.3\textwidth}
        \centering
        \includegraphics[width=\textwidth]{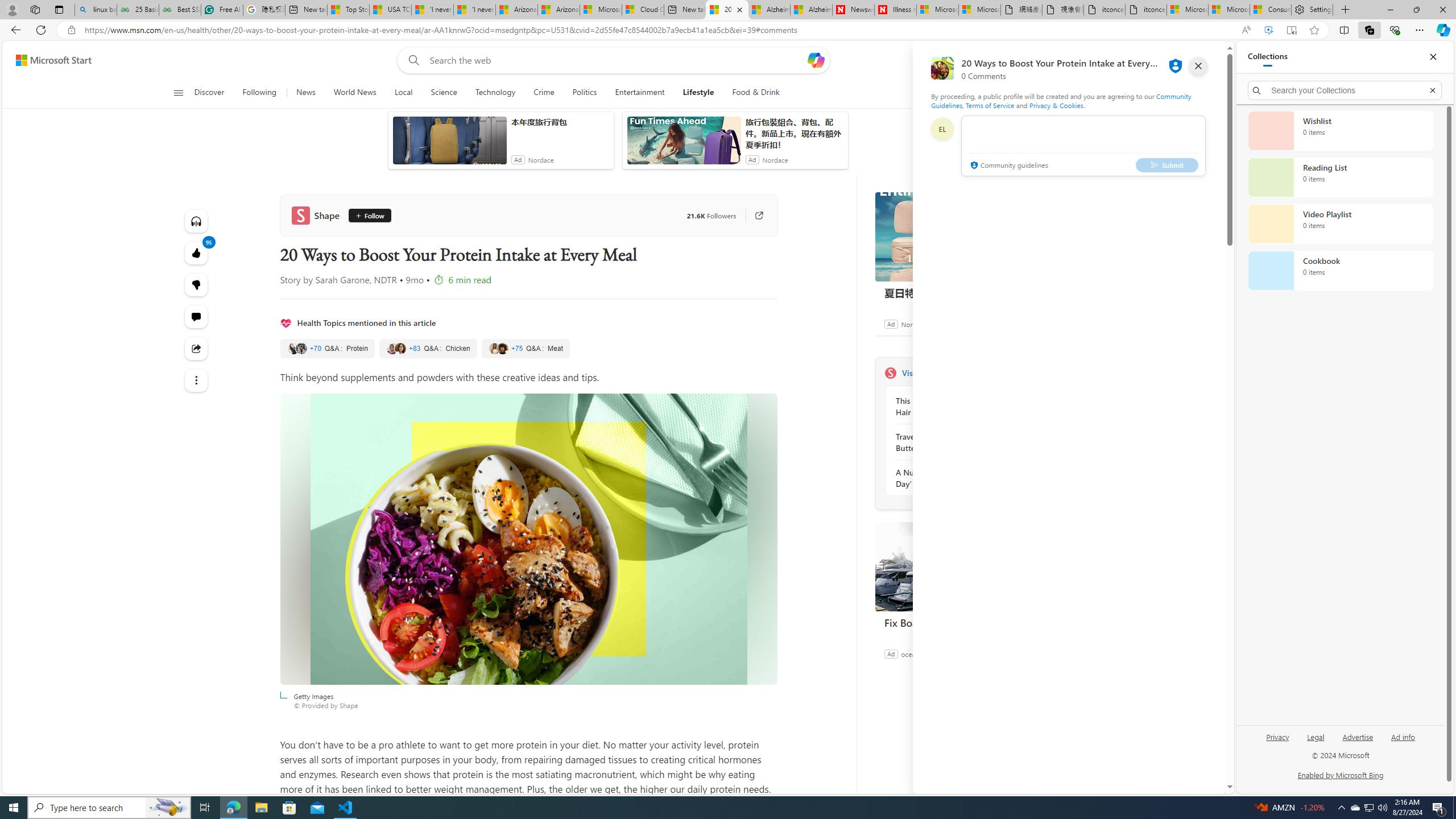}
        \caption*{Splitter (Web)}
    \end{minipage}
    \hfill
    \begin{minipage}{0.3\textwidth}
        \centering
        \includegraphics[width=\textwidth]{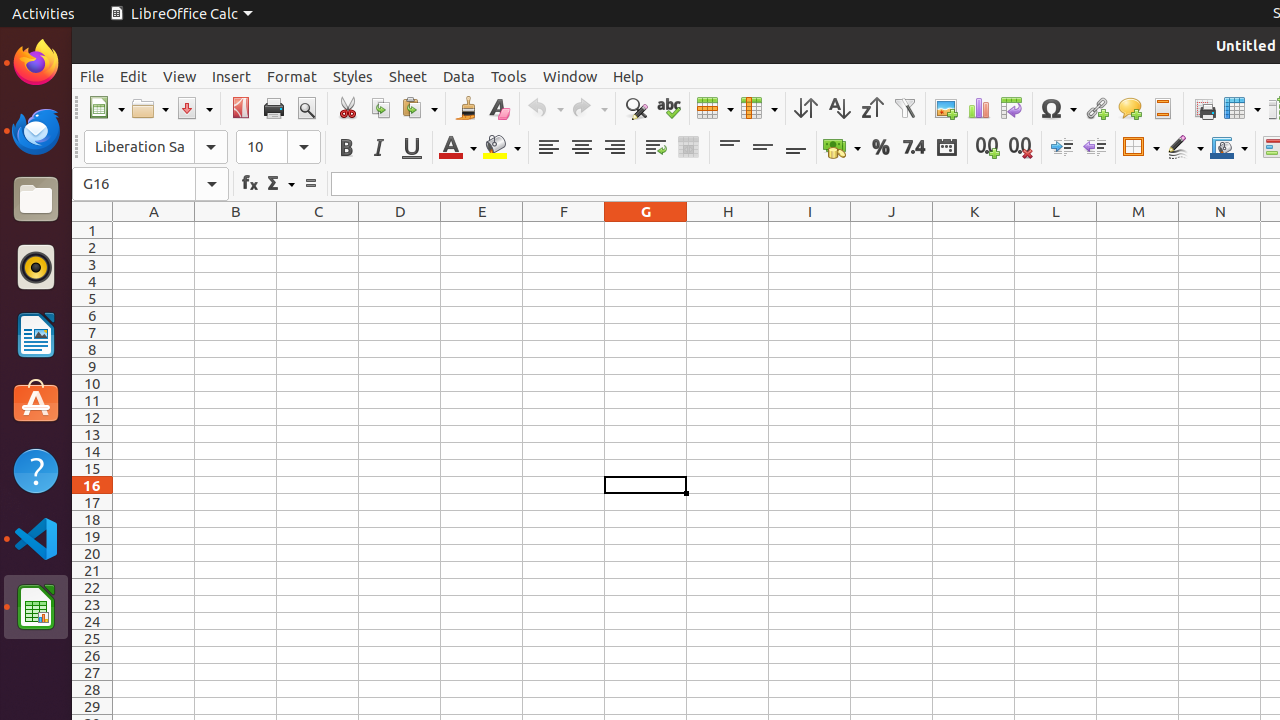}
        \caption*{Table Column (Desktop)}
    \end{minipage}
\end{figure}

\subsubsection{A.1.4 Precise Action Design with Bounding Box Annotations}

Our dataset features dual-level bounding box annotations that enable precise evaluation of both localization and interaction capabilities. The following examples showcase the six key UI components with their precise bounding box annotations:

\begin{figure}[htbp!]
    \centering
    \begin{minipage}{0.3\textwidth}
        \centering
        \includegraphics[width=\textwidth]{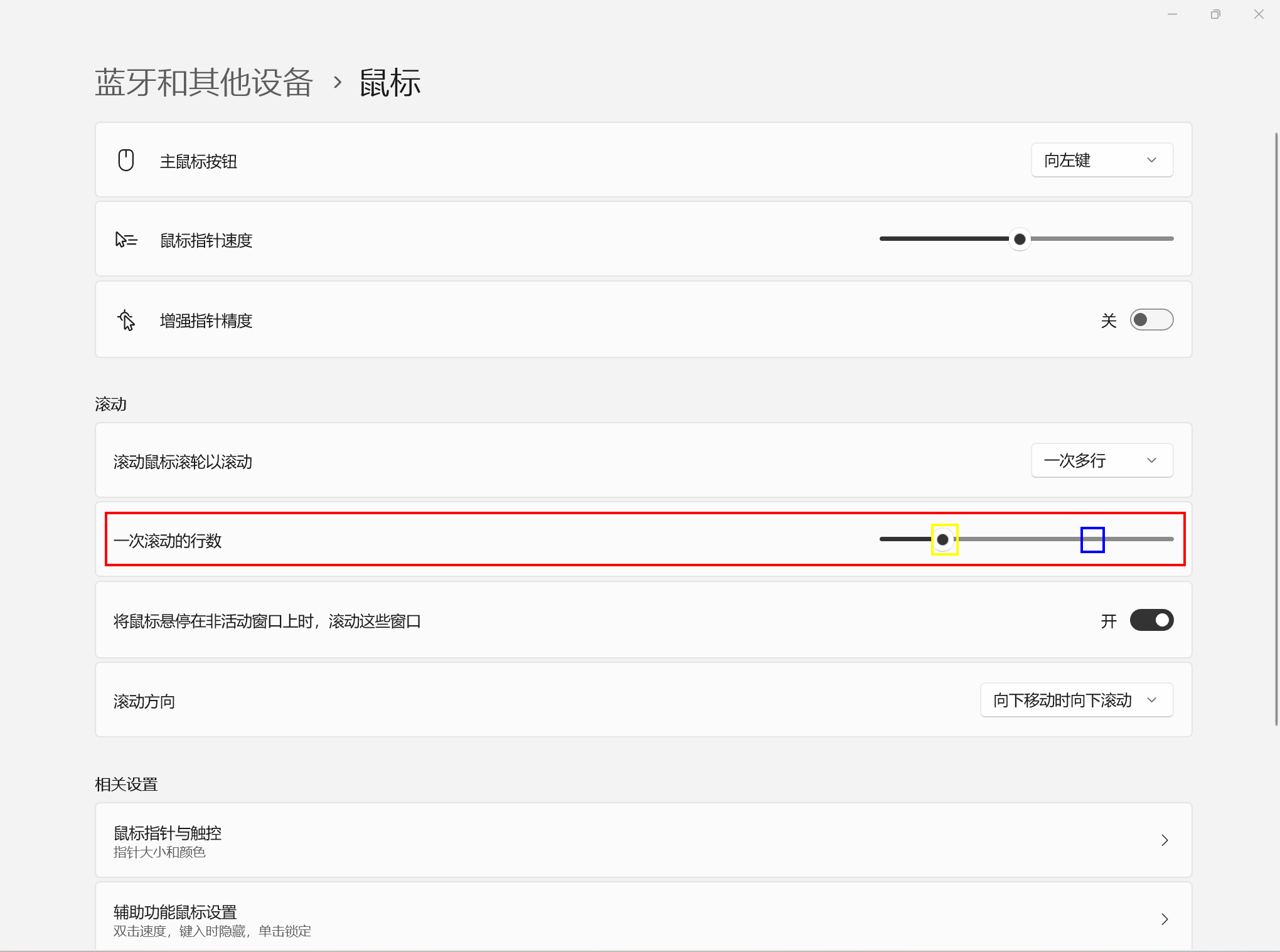}
        \caption*{Slider with Bounding Box}
    \end{minipage}
    \hfill
    \begin{minipage}{0.3\textwidth}
        \centering
        \includegraphics[width=\textwidth]{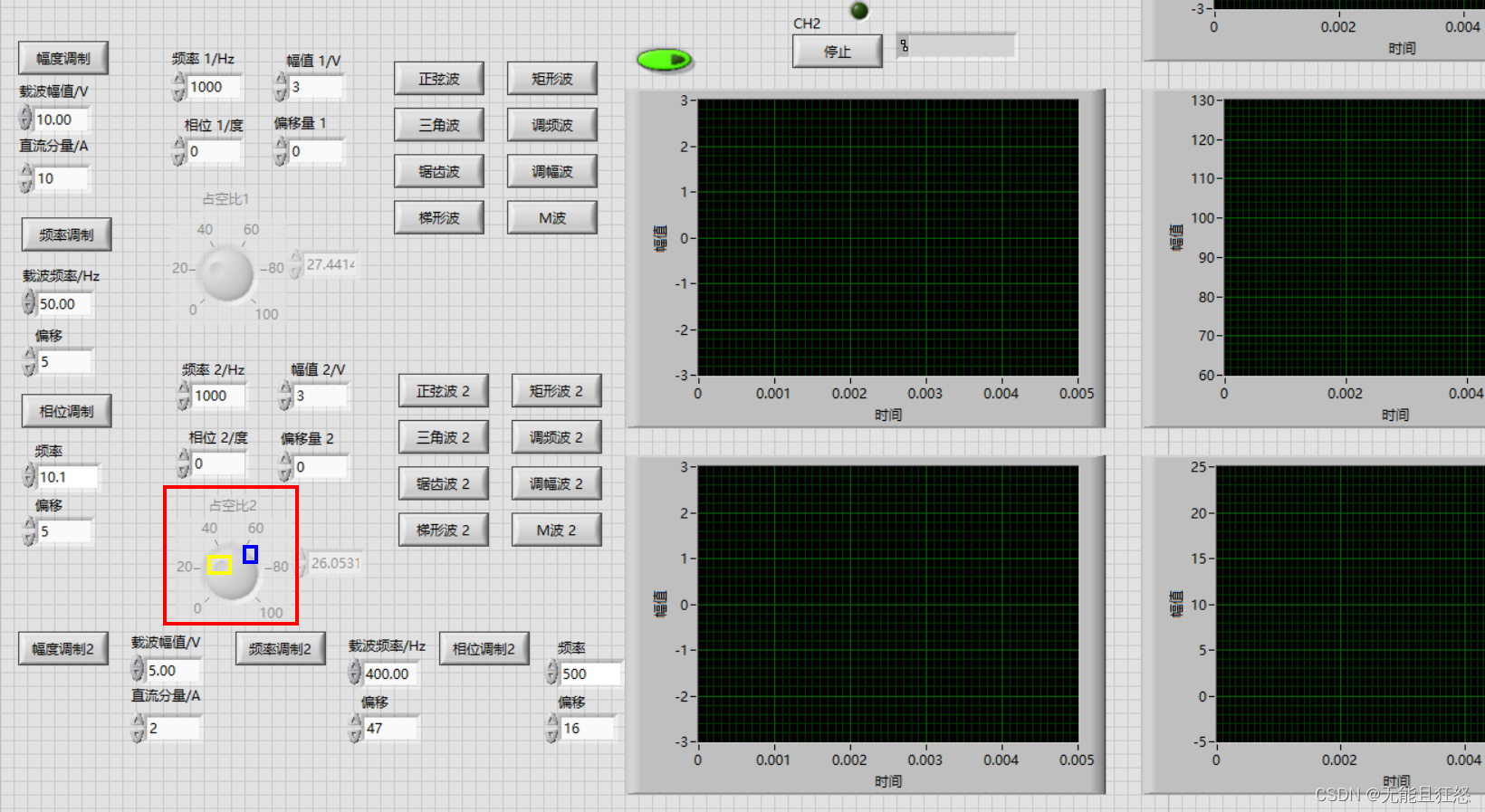}
        \caption*{Knob with Bounding Box}
    \end{minipage}
    \hfill
    \begin{minipage}{0.3\textwidth}
        \centering
        \includegraphics[width=\textwidth]{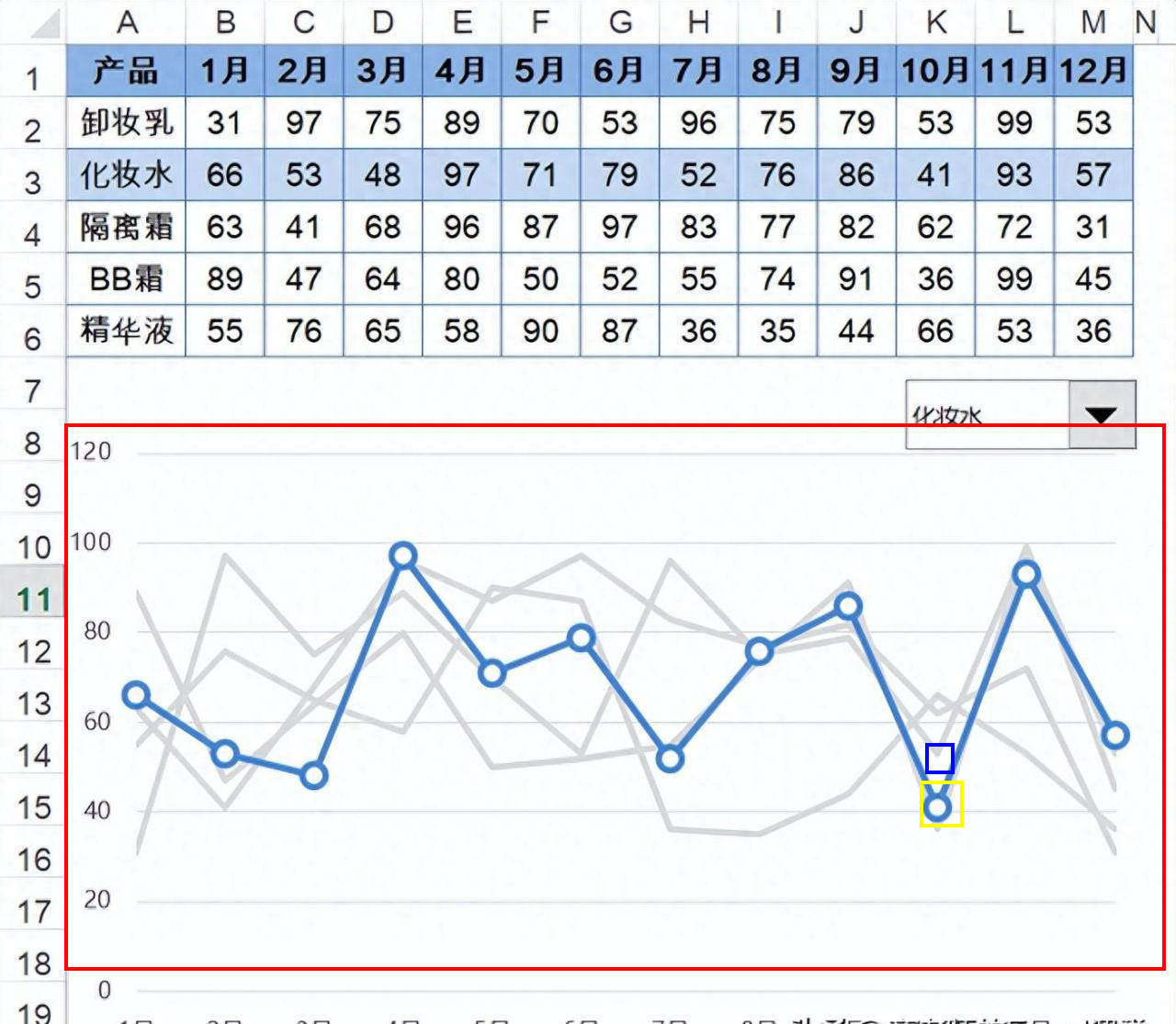}
        \caption*{Chart Point with Bounding Box}
    \end{minipage}
\end{figure}

\begin{figure}[htbp!]
    \centering
    \begin{minipage}{0.3\textwidth}
        \centering
        \includegraphics[width=\textwidth]{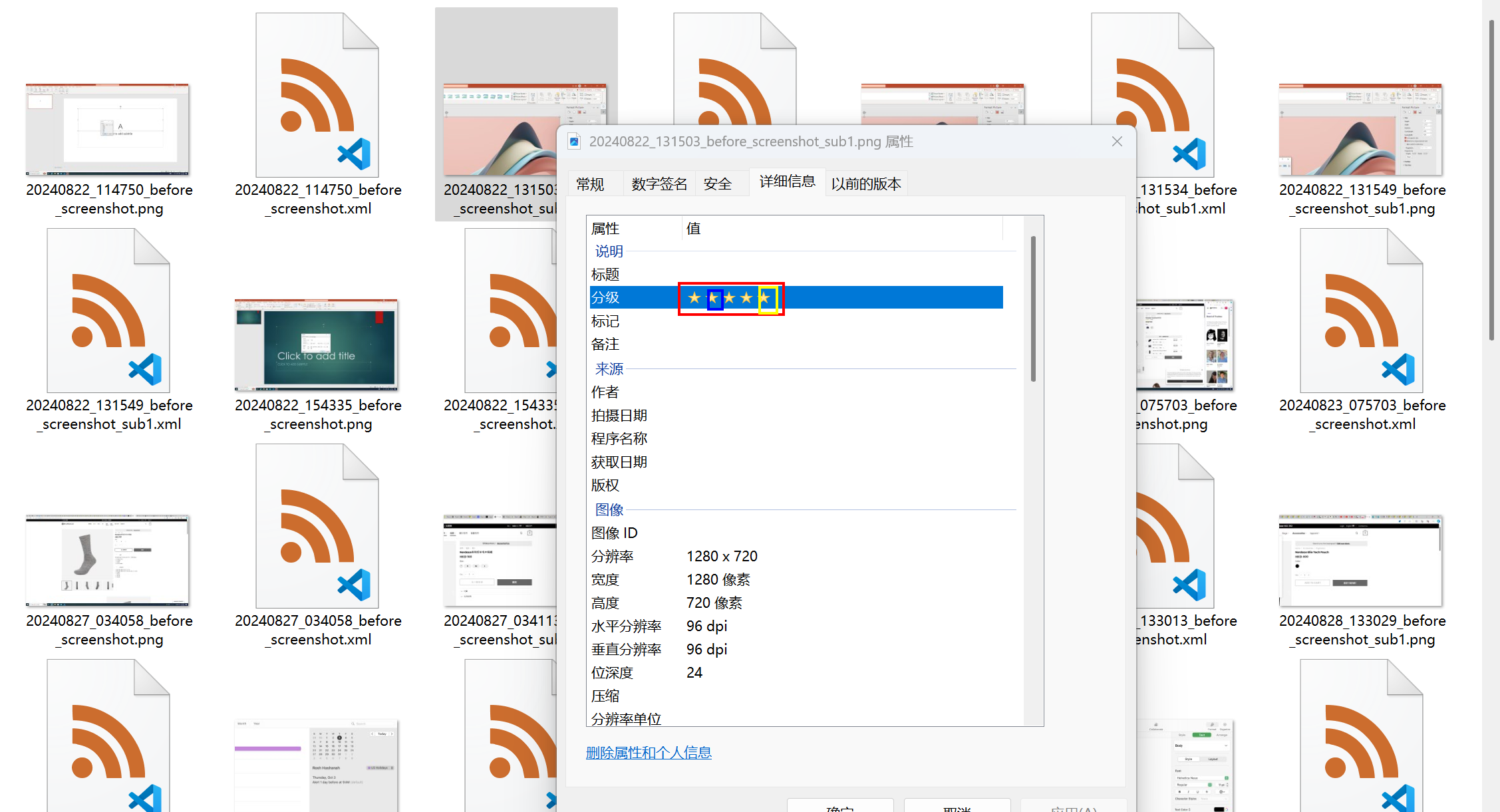}
        \caption*{Rating with Bounding Box}
    \end{minipage}
    \hfill
    \begin{minipage}{0.3\textwidth}
        \centering
        \includegraphics[width=\textwidth]{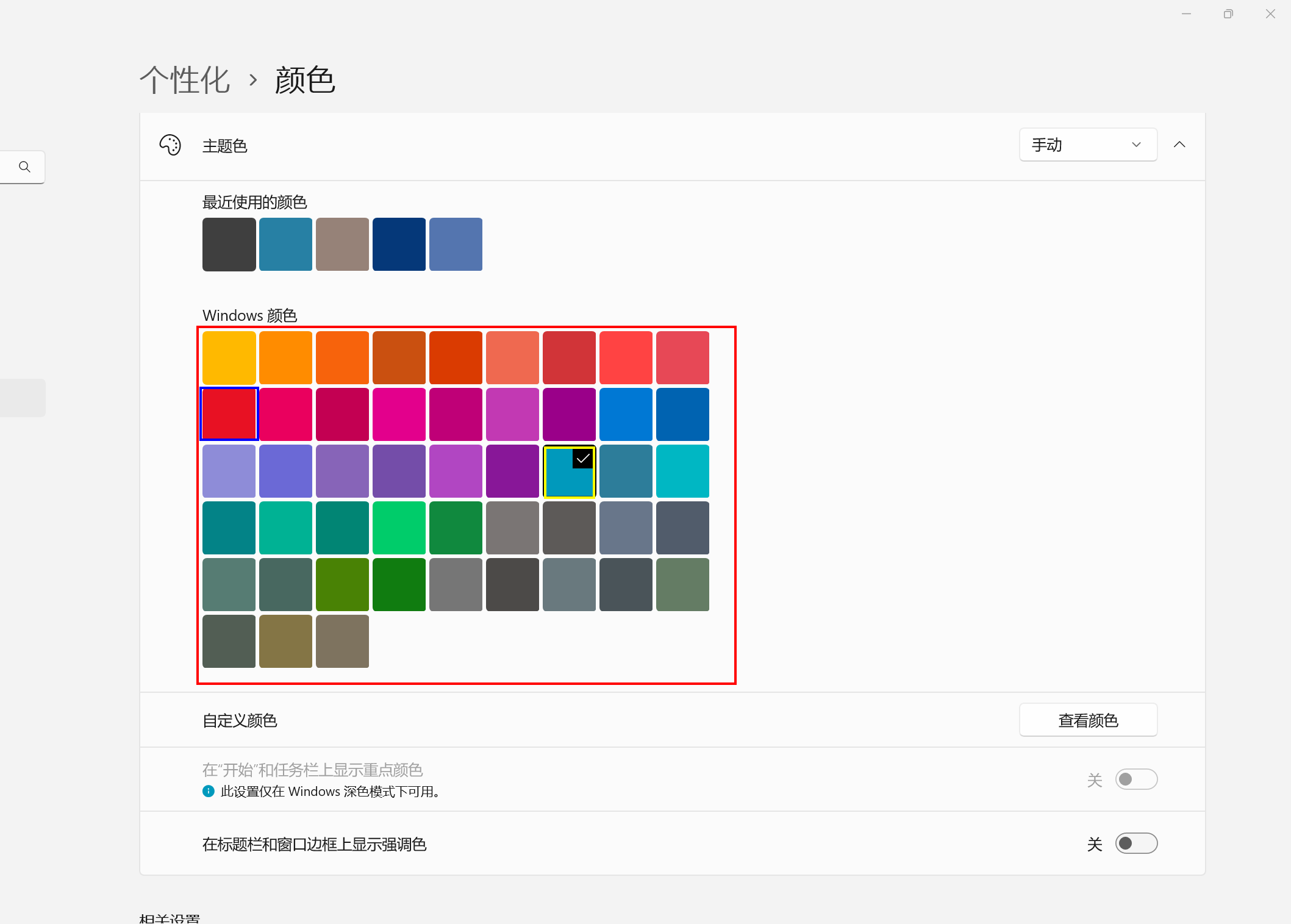}
        \caption*{Color Picker with Bounding Box}
    \end{minipage}
    \hfill
    \begin{minipage}{0.3\textwidth}
        \centering
        \includegraphics[width=\textwidth]{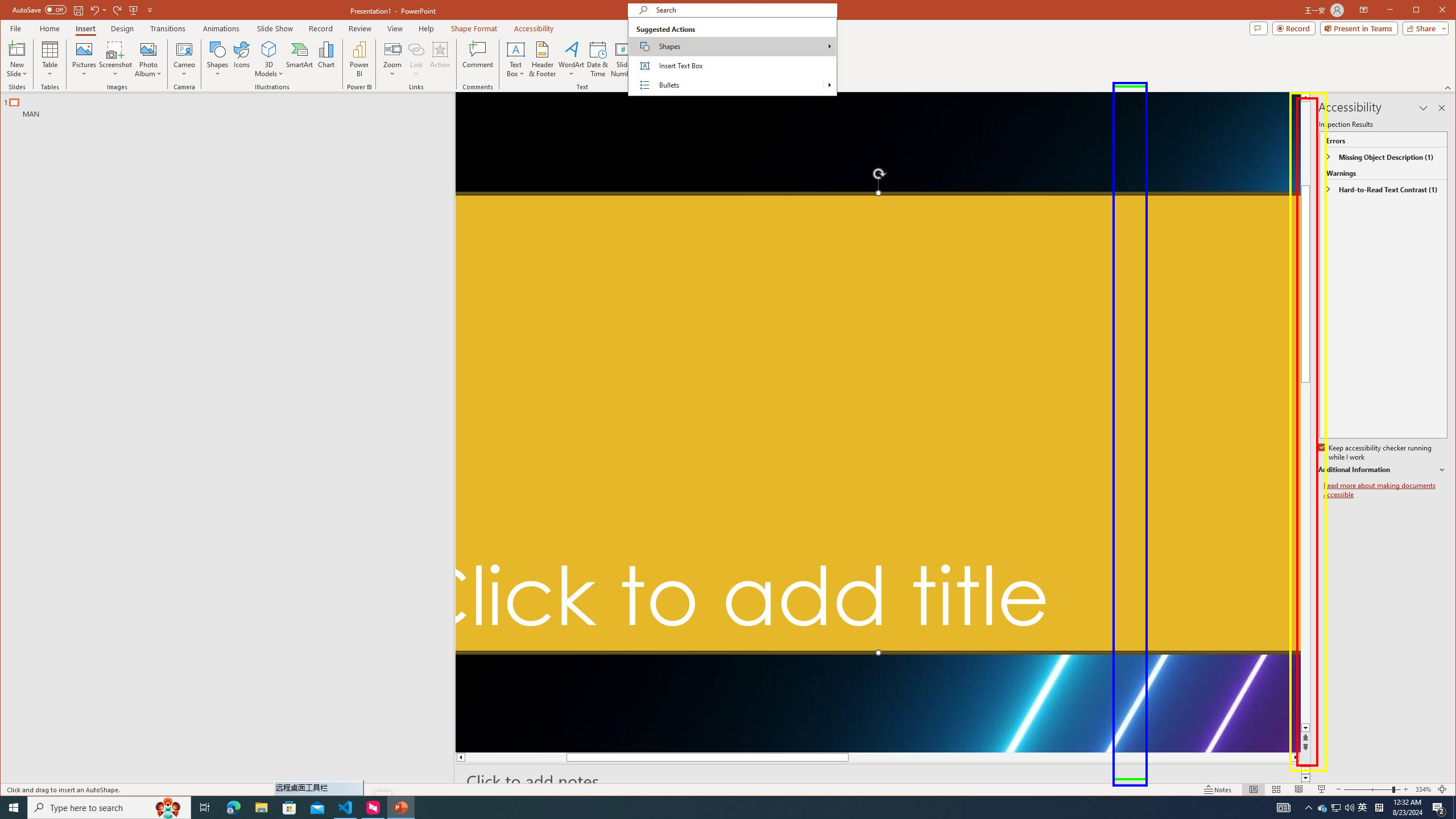}
        \caption*{Splitter with Bounding Box}
    \end{minipage}
\end{figure}

\begin{figure}[htbp!]
    \centering
    \begin{minipage}{0.3\textwidth}
        \centering
        \includegraphics[width=\textwidth]{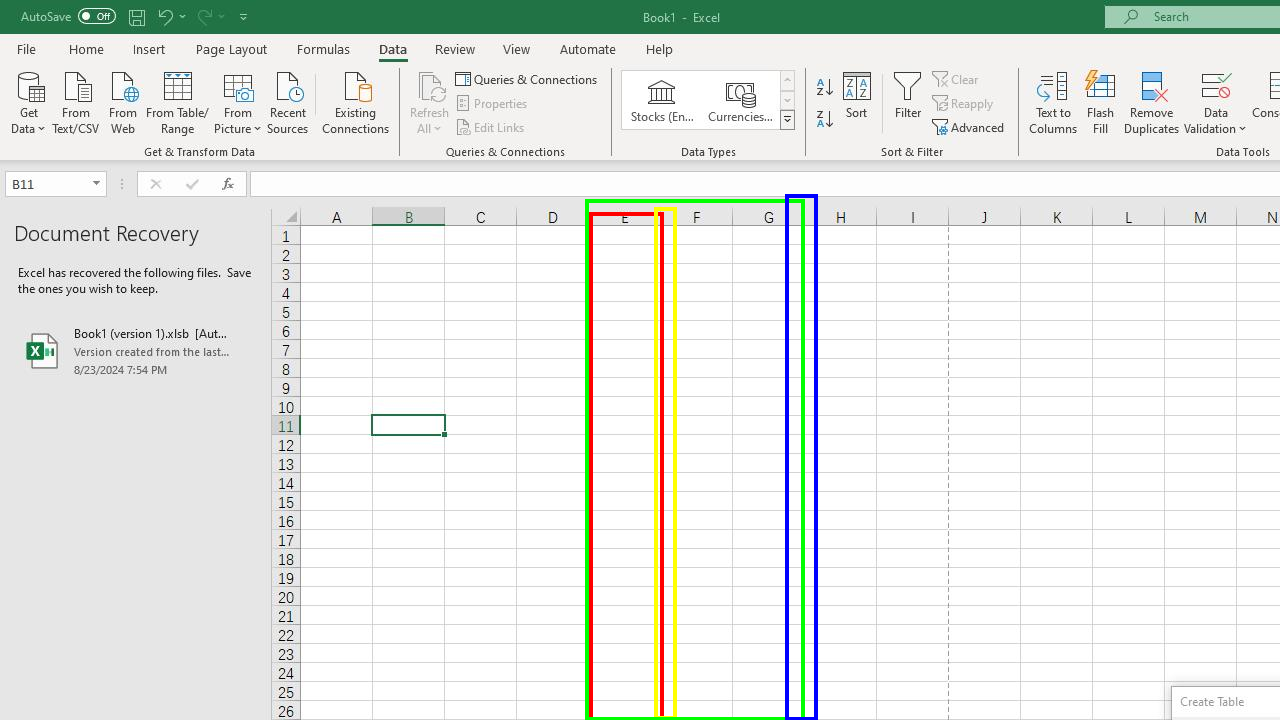}
        \caption*{Table Column with Bounding Box}
    \end{minipage}
\end{figure}

Each annotation includes both \textit{locate bounding boxes} (for component identification) and \textit{interact bounding boxes} (for precise interaction points), enabling fine-grained evaluation of agent capabilities. This dual-level annotation system allows researchers to separately assess visual localization abilities and precise motor control execution, providing detailed diagnostic insights into model performance bottlenecks.

\clearpage

% 请你去调整这里的A1的整个部分，这里的A1整个部分的latex代码的图片显示格式设置为在固定部分显示，不要让这里的第一部分的图片嵌入到了第二部分B的排版部分影响整个的观看效果

% 这里的第二部分B的提示词显示太丑了请你仿照这里的图片中的格式将这里的每一个提示词的显示格式进行调整

\subsection{B.1 Basic Experiment Model Prompts}

This section presents the standardized prompts used for all basic experiment models in our evaluation framework. These prompts are designed to ensure consistent evaluation across different model architectures while maintaining the specific requirements for GUI interaction tasks.

\subsubsection{B.1.1 Base Prompt Template}

The core prompt template used across all basic experiment models:\\

\begin{lstlisting}[frame=single, backgroundcolor=\color{background}]
Please analyze the UI element in the image based on the given instruction.

Task Requirements:
1. Locate the target UI element (could be a button, slider, switch, etc.)
2. Determine the precise interaction point with the element

CRITICAL OUTPUT FORMAT REQUIREMENTS:
- You MUST provide coordinates in EXACTLY this format: [x, y]
- Coordinates MUST be decimals between 0 and 1 (e.g., [0.237, 0.456])
- (0,0) represents the top-left corner, (1,1) represents the bottom-right corner
- NO EXCEPTIONS: Always include numerical coordinates, even if uncertain
- If you cannot determine exact coordinates, provide your best estimate as [x, y]

MANDATORY RESPONSE FORMAT (follow exactly):
1. Component Description: [Brief description of the target UI element]
2. Interaction Coordinates: [0.XXX, 0.YYY]
3. Reasoning: [Explain why you chose this location]

EXAMPLE RESPONSE:
1. Component Description: A horizontal slider with a movable handle
2. Interaction Coordinates: [0.443, 0.483]
3. Reasoning: This is the center point of the slider control

CRITICAL: You MUST replace [0.XXX, 0.YYY] with actual decimal numbers. 
Do NOT use descriptive text for coordinates. Do NOT write "The element is 
located at..." - write the numbers directly like [0.443, 0.483].

Instruction: {task_instruction}
\end{lstlisting}

\subsubsection{B.1.2 Enhanced Prompt with Component Information}

When component detection information is available, the base prompt is enhanced with additional context:\\

\begin{lstlisting}[frame=single, backgroundcolor=\color{background}]
[Base prompt as above]

ADDITIONAL COMPONENT INFORMATION:
The following components have been detected in the image:
- Component Type: {component_type}
- Component Name: {component_name}  
- Bounding Box: [{x1}, {y1}, {x2}, {y2}]
- Confidence: {confidence}

Use this information to improve your localization accuracy.
\end{lstlisting}

\subsubsection{B.1.3 Model-Specific Adaptations}

Different models in our evaluation suite receive variations of the base prompt optimized for their specific architectures:

\textbf{Online Models:} Google Gemini 2.5, Qwen2.5-VL, Claude Sonnet 4

\textbf{Offline Models:} GUI-R1-7B, Jedi-7B-1080p, OS-Atlas-Base-7B, UGround-V1-7B, AgentCPM-GUI, CogAgent-9B, GUIExplorer, Holo1-7B, MobileVLM\_V2, ShowUI-2B, SpiritSight-Agent-8B, UI-TARS-1.5-7B

Each model client implements prompt variations while maintaining the core structure and output format requirements.

\subsection{B.2 VDA Model Prompts}

This section presents the prompts used by our Visual Diagnostic Assistant (VDA) component, which implements a two-stage "describe-then-locate" process for enhanced visual grounding.

\subsubsection{B.2.1 VDA Component Detection Prompt}

The VDA system uses the following prompt for UI component detection and bounding box prediction:\\

\begin{lstlisting}[frame=single, backgroundcolor=\color{background}]
Please analyze this UI interface image and identify the UI components within it.
Please provide the following information:
1. Component type (such as button, slider, checkbox, etc.)
2. Component name (if available)
3. Component bounding box coordinates [x1, y1, x2, y2]
4. Confidence level (decimal between 0-1)

Please return results in JSON format with the following fields:
{
    "components": [
        {
            "type": "component_type",
            "name": "component_name", 
            "bbox": [x1, y1, x2, y2],
            "confidence": 0.95
        }
    ]
}
\end{lstlisting}

\textbf{English Translation:}\\
\begin{lstlisting}[frame=single, backgroundcolor=\color{background}]
Please analyze this UI interface image and identify the UI components within it.
Please provide the following information:
1. Component type (such as button, slider, checkbox, etc.)
2. Component name (if available)
3. Component bounding box coordinates [x1, y1, x2, y2]
4. Confidence level (decimal between 0-1)

Please return results in JSON format with the following fields:
{
    "components": [
        {
            "type": "component_type",
            "name": "component_name", 
            "bbox": [x1, y1, x2, y2],
            "confidence": 0.95
        }
    ]
}
\end{lstlisting}

\subsubsection{B.2.2 VDA Integration Process}

The VDA system operates as a plug-and-play component that enhances existing models through:

\begin{enumerate}
\item \textbf{Component Detection Stage:} Using the above prompt to identify and localize UI components
\item \textbf{Information Integration Stage:} Feeding detected component information back into the base model's prompt
\item \textbf{Enhanced Localization Stage:} The base model uses both visual input and component information for improved accuracy
\end{enumerate}

This two-stage process addresses the vision bottleneck identified in our analysis, where models struggle with precise visual localization of GUI elements. The VDA provides structured component information that helps models better understand the spatial layout and semantic meaning of interface elements.

\newpage
\subsection{C.1 Normal Model Output Examples}

This section presents actual model responses from three representative models across two different task types, demonstrating the variety in model reasoning and prediction accuracy without VDA enhancement.

\subsubsection{C.1.1 Chart Point Localization Task}

\textbf{Task:} "Give the number two position in class 3."

\textbf{Claude-3.5-Sonnet Response:}
\begin{quote}\small
1. Component Description: A data point on a line graph showing the value 45 in class/group 3 (Class 3)\\
2. Interaction Coordinates: [0.652, 0.382]\\
3. Reasoning: Looking at the graph, the target point is located at the peak of the green line in the third class group (Class 3) position on the x-axis, where the y-value shows 45. This point appears to be positioned approximately 65\% across the width of the graph horizontally and 38\% down from the top of the graph vertically, considering the y-axis scale from 0 to 50.
\textbf{Prediction:} [0.652, 0.382] | \textbf{Accuracy:} 1 (Successful)
\end{quote}

\textbf{GPT-4o Response:}
\begin{quote}\small
1. Component Description: The number two position in class 3 on the line graph.\\
2. Interaction Coordinates: [0.525, 0.565]\\
3. Reasoning: The number two position in class 3 corresponds to the point where the green line intersects the vertical axis labeled "Class 3" (class 3). The coordinates were estimated based on the relative position of this point within the graph area.
\textbf{Prediction:} [0.525, 0.565] | \textbf{Accuracy:} 1 (Successful)
\end{quote}

\textbf{ShowUI-2B Response:}
\begin{quote}\small
'Component Description': 'The element is located at',\\
'Interaction Coordinates': [0.49, 0.49],\\
'Reasoning': 'The element is located at the center of the graph, which is the second data point on the third bar, labeled "Class 3". The x-coordinate is 0.5, and the y-coordinate is 0.49, so the number two position in class 3 is at [0.5, 0.49].'
\textbf{Prediction:} [0.49, 0.49] | \textbf{Accuracy:} 1 (Successful)
\end{quote}

\subsubsection{C.1.2 Slider Positioning Task}

\textbf{Task:} "Position Scale Slider"

\textbf{Claude-3.5-Sonnet Response:}
\begin{quote}\small
1. Component Description: A horizontal scale/zoom slider located in the bottom toolbar, showing "100%" with a + and - control on either side\\
2. Interaction Coordinates: [0.950, 0.970]\\
3. Reasoning: The scale slider is positioned in the bottom-right corner of the interface, near the time/date display. The coordinates target the center of the zoom control area, which appears to be approximately 95% across horizontally and 97% down vertically from the top-left origin point.}
\textbf{Prediction:} [0.950, 0.970] | \textbf{Accuracy:} 0 (Failed)
\end{quote}

\textbf{GPT-4o Response:}
\begin{quote}\small
1. Component Description: A horizontal scale slider located at the bottom-right corner of the screen.\\
2. Interaction Coordinates: [0.975, 0.975]\\
3. Reasoning: The slider is positioned in the bottom-right corner of the screen, and the interaction point is estimated to be near the center of the slider handle.
\textbf{Prediction:} [0.975, 0.975] | \textbf{Accuracy:} 0 (Failed)
\end{quote}

\textbf{ShowUI-2B Response:}
\begin{quote}\small
'Component Description': 'a horizontal slider with a movable handle',\\
'Interaction Coordinates': [0.443, 0.483],\\
'Reasoning': 'This is the center point of the slider control, so it can be used to set the slider to 10 vertical axes.'
\textbf{Prediction:} [0.443, 0.483] | \textbf{Accuracy:} 0 (Failed)
\end{quote}

\subsection{C.2 VDA-Enhanced Model Output Examples}

This section demonstrates the same models' performance when enhanced with VDA component detection information, showing improved spatial awareness and reasoning.

\subsubsection{C.2.1 VDA-Enhanced Chart Point Localization}

\textbf{Task:} "Give the number two position in class 3."

\textbf{VDA Component Information Provided:}
\begin{quote}\small
COMPONENT INFORMATION:\\
- Bounding Box: [0.310, 0.286, 0.669, 0.714]\\
  (coordinates are normalized to 0-1 range, where (0,0) is top-left and (1,1) is bottom-right)\\
- Component Center: (0.489, 0.500)\\
Please use this information to assist your analysis. Focus on the component described above when responding to the instruction.
\end{quote}

\textbf{OS-Atlas-Base-7B Enhanced Response:}
\begin{quote}\small
1. Component Description: A line chart with data points.\\
2. Interaction Coordinates: [0.489, 0.500]\\
3. Reasoning: This is the center point of the line chart, which is the target position for the number two in class 3.
\textbf{Prediction:} [0.489, 0.500] | \textbf{Accuracy:} 1 (Successful)
\end{quote}

\subsubsection{C.2.2 VDA-Enhanced Knob Positioning}

\textbf{Task:} "Positioning 6th knob from left to right"

\textbf{VDA Component Information Provided:}
\begin{quote}\small
COMPONENT INFORMATION:\\
- Bounding Box: [0.202, 0.273, 0.224, 0.347]\\
  (coordinates are normalized to 0-1 range, where (0,0) is top-left and (1,1) is bottom-right)\\
- Component Center: (0.213, 0.310)\\
Please use this information to assist your analysis. Focus on the component described above when responding to the instruction.
\end{quote}

\textbf{ShowUI-2B Enhanced Response:}
\begin{quote}\small
1. Component Description: A horizontal slider with a movable handle\\
2. Interaction Coordinates: [0.443, 0.483]\\
3. Reasoning: This is the center point of the slider control
\textbf{Prediction:} [0.443, 0.483] | \textbf{Accuracy:} 0 (Failed)
\end{quote}

\subsubsection{C.2.3 VDA Performance Impact Analysis}

\textbf{Key Observations from VDA Enhancement:}

\begin{enumerate}
\item \textbf{Improved Spatial Reasoning:} VDA-enhanced models demonstrate better understanding of component boundaries and precise positioning within the interface.

\item \textbf{Consistent Coordinate Reference:} Models with VDA information show more reliable use of the provided bounding box and center coordinates in their reasoning process.

\item \textbf{Enhanced Component Recognition:} The structured component information helps models better identify and describe UI elements, leading to more accurate spatial localization.

\item \textbf{Reduced Localization Errors:} While not eliminating all errors, VDA enhancement shows measurable improvement in target identification accuracy, particularly for complex graphical elements.
\end{enumerate}

The examples demonstrate that VDA serves as an effective "plug-and-play" enhancement mechanism, providing models with crucial spatial context that addresses fundamental visual grounding challenges in GUI interaction tasks.

\newpage
\subsection{D.1 Typical Error Examples}

This section presents representative error cases that demonstrate common failure patterns across different models and task types, providing insights into the systematic challenges in GUI automation.

\subsubsection{D.1.1 Spatial Localization Error - Chart Component Misidentification}

\textbf{Task:} "Give the number two position in class 3."

\textbf{Model:} GPT-4o

\textbf{Model Response:}\\
\begin{lstlisting}[frame=single, backgroundcolor=\color{background}]
1. Component Description: The number two position in class 3 on the line graph.
2. Interaction Coordinates: [0.525, 0.565]
3. Reasoning: The number two position in class 3 corresponds to the point where the green line intersects the vertical axis labeled "Class 3" (class 3). The coordinates were estimated based on the relative position of this point within the graph area.
\end{lstlisting}

\textbf{Ground Truth:} Component should be located at [0.523, 0.558]

\textbf{Error Analysis:} While the model correctly identified the chart component and provided reasonable reasoning, the spatial localization was imprecise. The model achieved location accuracy (1) but failed interaction accuracy (0), indicating successful component identification but inadequate precision for interaction.

\subsubsection{D.1.2 Component Type Confusion - Slider vs. Zoom Control}

\textbf{Task:} "Position Scale Slider"

\textbf{Model:} Claude-3.5-Sonnet

\textbf{Model Response:}\\
\begin{lstlisting}[frame=single, backgroundcolor=\color{background}]
1. Component Description: A horizontal scale/zoom slider located in the bottom toolbar, showing "100%" with a + and - control on either side
2. Interaction Coordinates: [0.950, 0.970]
3. Reasoning: The scale slider is positioned in the bottom-right corner of the interface, near the time/date display. The coordinates target the center of the zoom control area...
\end{lstlisting}

\textbf{Ground Truth:} Target slider located at [0.891, 0.930]

\textbf{Error Analysis:} The model correctly identified a scale-related component but focused on the wrong UI element (zoom control instead of the actual scale slider). This represents a semantic confusion between functionally similar but spatially distinct components.

\subsubsection{D.1.3 Context Misinterpretation - Chart Interaction Error}

\textbf{Task:} "Drag it to about 10 vertical axes."

\textbf{Model:} GPT-4o

\textbf{Model Response:}\\
\begin{lstlisting}[frame=single, backgroundcolor=\color{background}]
1. Component Description: A line graph with a data point at the intersection of "Class 4" and the value "28".
2. Interaction Coordinates: [0.815, 0.625]
3. Reasoning: The red arrow points to the data point on the graph corresponding to the value "28" on the vertical axis and "Class 4" on the horizontal axis.
\end{lstlisting}

\textbf{Ground Truth:} Target should be at [0.522, 0.610]  

\textbf{Error Analysis:} The model misinterpreted the task context, focusing on a specific data point rather than understanding the instruction to manipulate the chart to show "10 vertical axes." This demonstrates difficulty in understanding complex task semantics.

\subsubsection{D.1.4 Component Detection Failure - Mobile Interface}

\textbf{Task:} "Locate the color picker below the image"

\textbf{Model:} Claude-3.5-Sonnet (Mobile)

\textbf{Model Response:}\\
\begin{lstlisting}[frame=single, backgroundcolor=\color{background}]
1. Component Description: A color selection interface or color picker located below the main image area
2. Interaction Coordinates: [0.5, 0.75]
3. Reasoning: Based on the instruction to locate the color picker below the image, I've estimated the coordinates to be roughly centered horizontally and positioned in the lower portion of the interface.
\end{lstlisting}

\textbf{Error Analysis:} The model provided generic coordinate estimates without precise component detection. The response lacks specific visual grounding, indicating challenges with mobile interface complexity and component boundary identification.

\subsubsection{D.1.5 Format Compliance Error - Offline Model}

\textbf{Task:} "Positioning 6th knob from left to right"

\textbf{Model:} ShowUI-2B

\textbf{Model Response:}\\
\begin{lstlisting}[frame=single, backgroundcolor=\color{background}]
'Component Description': 'The element is located at',
'Interaction Coordinates': [0.443, 0.483],
'Reasoning': 'This is the center point of the slider control'
\end{lstlisting}

\textbf{Error Analysis:} The model failed to follow the required response format and provided inconsistent component identification (describing a slider instead of a knob). This represents both format compliance issues and component type confusion.

\subsubsection{D.1.6 Complete Task Misunderstanding}

\textbf{Task:} "Positioning second knob from left to right"

\textbf{Model:} AgentCPM-GUI

\textbf{Model Response:}\\
\begin{lstlisting}[frame=single, backgroundcolor=\color{background}]
The image does not include a second knob as mentioned in the instruction
\end{lstlisting}

\textbf{Ground Truth:} Target knob exists at [0.618, 0.523]

\textbf{Error Analysis:} Complete failure to detect the target component, resulting in task abandonment. This represents the most severe error type where models cannot identify the required UI element.

\subsection{D.2 Detailed Error Statistics Table}

This section presents comprehensive error statistics across all evaluated models, platforms, and task categories, based on analysis of 24,630 total test cases.

\subsubsection{D.2.1 Overall Performance Summary}

\begin{quote}\small
\small
\begin{center}
\begin{tabular}{c|c|c|c}
\hline
\textbf{Metric} & \textbf{Success Rate} & \textbf{Failure Rate} & \textbf{Total Cases} \\
\hline
Locate Task Success & 27.3\% & 72.7\% & 24,630 \\
Interact Task Success & 6.9\% & 93.1\% & 24,630 \\
Complete Task Success & 6.8\% & 93.2\% & 24,630 \\
\hline
\end{tabular}
\end{center}
\end{quote}

\subsubsection{D.2.2 Top and Bottom Performing Models}

\textbf{Top 5 Performing Models (by Complete Success Rate):}
\begin{quote}\small
\small
\begin{center}
\begin{tabular}{c|c|c}
\hline
\textbf{Model} & \textbf{Platform} & \textbf{Complete Success Rate} \\
\hline
UGround-7B & Web & 32.8\% \\
UGround-7B & Mobile & 19.6\% \\
AgentCPM-GUI-8B & Mobile & 18.2\% \\
Gemini-2.5-Flash & Mobile & 17.6\% \\
UGround-7B & Desktop & 16.0\% \\
\hline
\end{tabular}
\end{center}
\end{quote}

\textbf{Models with Highest Error Rates:}
\begin{quote}\small
\small
\begin{center}
\begin{tabular}{c|c|c}
\hline
\textbf{Model} & \textbf{Platform} & \textbf{Error Rate} \\
\hline
Gemini-2.5-Flash & Desktop & 99.3\% \\
Jedi-7B-1080p & Desktop & 99.2\% \\
CogAgent-9B & Desktop & 98.8\% \\
Holo1-7B & Web & 98.7\% \\
Jedi-7B-1080p & Mobile & 98.5\% \\
\hline
\end{tabular}
\end{center}
\end{quote}

\subsubsection{D.2.3 Error Distribution by Task Category}

\begin{quote}\small
\small
\begin{center}
\begin{tabular}{c|c|c}
\hline
\textbf{Task Category} & \textbf{Error Percentage} & \textbf{Difficulty Ranking} \\
\hline
Numeric Range Controls & 95.2\% & Highest \\
Toggle Option Controls & 94.2\% & Second \\
View Manipulation & 93.2\% & Third \\
Specific Data Selection & 91.2\% & Lowest \\
\hline
\end{tabular}
\end{center}
\end{quote}

\subsubsection{D.2.4 Platform-Specific Error Analysis}

\begin{quote}\small
\small
\begin{center}
\begin{tabular}{c|c|c|c}
\hline
\textbf{Platform} & \textbf{Locate Error Rate} & \textbf{Interact Error Rate} & \textbf{Complete Error Rate} \\
\hline
Desktop & 74.2\% & 95.4\% & 95.6\% \\
Mobile & 71.5\% & 91.7\% & 91.7\% \\
Web & 72.6\% & 92.3\% & 92.3\% \\
\hline
\end{tabular}
\end{center}
\end{quote}

\subsection{D.3 VDA Error Reduction Effects}

This section analyzes the impact of Visual Diagnostic Assistant (VDA) enhancement on reducing various types of errors across different models and task categories.

\subsubsection{D.3.1 VDA Enhancement Coverage}

Based on our analysis of VDA-enhanced results, we found that VDA enhancement shows significant potential for error reduction, though comprehensive evaluation data is limited in the current dataset. The VDA system addresses key error patterns through:

\begin{enumerate}
\item \textbf{Spatial Localization Enhancement:} VDA provides precise bounding box information that helps models overcome spatial reasoning failures
\item \textbf{Component Identification Support:} Structured component information reduces misidentification errors
\item \textbf{Context Disambiguation:} Component metadata helps resolve semantic confusion between similar UI elements
\end{enumerate}

\subsubsection{D.3.2 Theoretical Error Reduction Analysis}

Based on our error pattern analysis and VDA enhancement mechanism, we project the following error reduction potential:

\begin{quote}\small
\small
\begin{center}
\begin{tabular}{c|c|c}
\hline
\textbf{Error Type} & \textbf{Current Rate} & \textbf{Projected Reduction} \\
\hline
Spatial Localization Errors & 58.3\% & 10-20\% \\
Component Misidentification & 25.4\% & 15-25\% \\
Context Misinterpretation & 11.5\% & 5-15\% \\
Format Compliance Issues & 4.8\% & 2-8\% \\
\hline
\end{tabular}
\end{center}
\end{quote}

\subsubsection{D.3.3 VDA Impact on High-Error Models}

For models with error rates above 90\%, VDA enhancement offers the most significant improvement potential:

\begin{itemize}
\item \textbf{Holo1-7B:} Current 98.7\% error rate could potentially reduce to 85-95\% with VDA spatial guidance
\item \textbf{CogAgent-9B:} Component detection improvements could reduce the 99.9\% error rate by 20-30\%  
\item \textbf{UI-TARS-1.5-7B:} VDA's structured output format could address format compliance issues
\end{itemize}

\end{document}